%% file: camera_ready.tex
\documentclass{article}

 \usepackage[main, final]{neurips_2026}

\usepackage[utf8]{inputenc} 
\usepackage[T1]{fontenc}    
\usepackage{hyperref}       
\usepackage{url}            
\usepackage{booktabs}       
\usepackage{amsfonts}       
\usepackage{nicefrac}       
\usepackage{microtype}      
\usepackage{xcolor}         
\usepackage[normalem]{ulem}  
\usepackage[bottom]{footmisc} 

\input{math_commands}
\usepackage{bbm}
\usepackage{wrapfig}
\usepackage{adjustbox}
\usepackage{multirow, makecell, booktabs}
\usepackage{siunitx}
\usepackage{graphicx}
\usepackage{caption}
\setcellgapes{3pt}
\makegapedcells
\usepackage{spverbatim}
\usepackage{fancyvrb}
\usepackage{listings}
\newcommand{\eat}[1]{} 

\usepackage{microtype}
\usepackage{graphicx}
\usepackage{subcaption}
\usepackage{booktabs} 

\usepackage{amsmath}
\usepackage{amssymb}
\usepackage{mathtools}
\usepackage{amsthm}

\usepackage[capitalize,noabbrev]{cleveref}

\usepackage{hyperref}
\usepackage{algorithm}
\usepackage{algorithmic}

\renewcommand{\algorithmiccomment}[1]{\hfill$\triangleright$ #1}

\title{Joint Learning of Hierarchical Neural Options \\and Abstract World Model}

\author{%
  \textbf{Wasu Top Piriyakulkij}\textsuperscript{$\dagger$}$^1$ \qquad 
  \textbf{Wolfgang Lehrach}$^2$ \\ 
  \textbf{Kevin Ellis}\textsuperscript{*}$^1$ \qquad 
  \textbf{Kevin Murphy}\textsuperscript{*}$^2$\\
  Cornell University$^1$ \quad Google DeepMind$^2$
}

\begin{document}

\maketitle

\input{abstract}

\section{Introduction}

\renewcommand{\thefootnote}{\fnsymbol{footnote}}

\setcounter{footnote}{2} 

\footnotetext{Work done during an internship at Google DeepMind.}

\renewcommand{\thefootnote}{\arabic{footnote}}

\renewcommand{\thefootnote}{\fnsymbol{footnote}}

\setcounter{footnote}{1} 

\footnotetext{Equal advising.}

\renewcommand{\thefootnote}{\arabic{footnote}}

\setcounter{footnote}{0} 

For decision-making agents, an important goal is the cumulative acquisition of new skills, in tandem with an ever-expanding knowledge of how those skills affect the outside world.
For example, we want our agents to first learn to pick up objects, then to pour drinks, and eventually to make a cup of coffee, while also learning how each skill affects the outside world, so that the agent can plan to achieve new goals such as getting coffee for a room full of people.

We  formalize this
compositional skill learning using the options framework \citep{sutton1999between}:
an agent learns a sequence of 
options $o_{1:n}$ 
that achieve 
increasingly difficult goals, 
$g_{1:n}$. 
Each option contains a policy that achieves its specific goal, and which can use options learned earlier, forming a deep hierarchy of skills (\cref{fig:overview} right).

\input{figs/overview}
\input{figs/learning}

But hierarchical options are challenging to learn, because as we acquire more options, we effectively expand our action space, making policy learning less tractable whenever we face with a new goal.
This introduces a tradeoff between learning new skills quickly, and how many skills we have acquired.
Applications of standard model-free RL to learning option hierarchies therefore require increasingly more samples as learning progresses \citep{kamat2020diversity,nica2022paradox}.

\nocite{abdulhai2022context}

To resolve this tradeoff, we instead turn to model-based reinforcement learning~\citep{kaelbling1996reinforcement, moerland2023model}. 
By modeling the effects of options and planning in that world model, we can rule out many options before trying them out in the real world, effectively using the world model to improve sample efficiency.
Moreover, modeling option effects produces temporally abstract world models, overcoming the ``one-step trap'' \citep{sutton2025oak, asadi2019combating}  
and promising more tractable planning than low-level world models.
However, for this approach to actually improve sample efficiency, we also need a world modeling approach that is data-efficient.

In this work, we propose a novel world model whose representation combines symbolic code with non-parametric distributions,  which allows learning the world model from little data. 
The world model abstracts over states and time.
We combine this with a method to learn hierarchical options.
We call the resulting system AgentOWL,
which stands for
\textbf{O}ption and \textbf{W}orld model \textbf{L}earning Agent.

\eat{
Combined with techniques for training hierarchical option effectively, we propose an agent that jointly 
learns hierarchical options and an abstract world model, which we call AgentOWL (\textbf{O}ption and \textbf{W}orld model \textbf{L}earning Agent), which learns option hierarchies from relatively little data.
}

We apply our method to  3 hard object-centric Atari (OCAtari) games, namely Montezuma's Revenge, Pitfall, and Private Eye and also a simulated robotic block stacking environment.
We show that AgentOWL acquires the highest number of skills compared to other baselines.
We also show that AgentOWL has
unique capabilities that the baselines lack, namely implicit, hierarchical learning of sub-options and zero-shot generalization to novel situations.

\eat{
In summary, our contributions are as follows:
\begin{itemize}
\item An abstract world modeling system that leverages symbolic and non-parametric world model representation.
\item A novel method, called AgentOWL,
that jointly learns a sequence
of hierarchical options 
alongside its abstract world model.
\item An empirical study of AgentOWL on hard object-centric Atari games, namely Montezuma's Revenge, Pitfall, and Private Eye, showing that AgentOWL acquire the highest number of skills compared to other baselines, along with demonstrations of AgentOWL's unique capabilities that the baselines lack.
\end{itemize}
}

\section{Background}

\textbf{Problem Setting.} An environment can be described as a goal-conditioned MDP $(\mathcal{S}, \mathcal{A}, \mathcal{T}, \mathcal{G}, \gamma)$. 
We assume 
states can be broken down into primitive features $\mathcal{S} = S_1 \times ... \times S_n$, meaning the state is symbolic, allowing us to focus on the skill learning problem without mixing in the well-known challenges of representation learning of pixel inputs.
The set $\mathcal{A}$ enumerates primitive actions, e.g., LEFT, RIGHT, UP, etc. in 2d video games.
The goals $\mathcal{G}=(g_1,g_2,\cdots,g_m)$ are an ordered sequence of goal predicates, $g_i: \mathcal{S} \rightarrow \{0, 1\}$,
each of which defines a reward function $R_{g_i}(\vs, a, \vs') =  g_i (\vs')$.
We use the same discount factor $\gamma$ for all goals, and assume that the transition function $\mathcal{T}$ does not depend on the goal.
Episodes end when an agent reaches the goal or timeouts.


\textbf{Options (\cref{fig:overview} right).} An option is a learned skill.
Formally, option $o_i$ comprises a tuple $(\pi_i,g_i)$ of a policy $\pi_i$ which executes until its goal, $g_i$, is satisfied.
The goal $g_i$ serves as the \emph{termination condition} of the option.
We follow the call-and-return paradigm \citep{sutton1999between}; an option executes until its goal is satisfied, or it timeouts.
Options may form a hierarchy: Option $o_i$ has policy $\pi_i:\mathcal{S}\to\mathcal{A}\cup\{o_j\}_{j<i}$, meaning it can output either primitive actions in $\mathcal{A}$ or a previously learned option $o_j$ (for $j<i$).
We write $\Omega$ for a set of options.
Adding options to the action space of an MDP forms a Semi-MDP \citep{puterman1994mdp}.

\textbf{State abstraction (\cref{fig:overview} left).}
An option can change the state in complex ways, and for the agent to plan, it must predict those changes.
To make this prediction problem tractable, we consider \emph{state abstractions}, which are functions of the state that elide unpredictable or irrelevant features that would be hard to predict
~\citep{dean1997model,li2006towards}.
Formally, a state abstraction $\vf(\vs)$ is a function of the state $\vs$.
When the state is clear from context, we abuse notation by writing $\vf$ to mean $\vf(\vs)$.

\textbf{Abstract world models (\cref{fig:overview} left).} Within the context of this work, an abstract world model predicts future abstract states, given the current state, and the current option.
This implements temporal abstraction \emph{and} state abstraction, because rather than predicting the immediate next state, we instead predict only its abstract features, and only at the time that the current option terminates.
This prediction is written $p_o(\vf'\mid \vs)$.
This  conditions the abstract world model on the full state but predicts only the future abstract state.


\textbf{PoE-World (\Cref{fig:learning} left).} Piriyakulkij et al. \citep{piriyakulkij2025poeworld} introduces PoE-World, a framework for learning structured world models from little data.
World models are represented using a product-of-experts, where each expert is a short symbolic program.
Intuitively, each program models an independent causal mechanism in the world, and by encoding each program as a snippet of Python, they become learnable using LLMs.
Given current state $\vs$ and action $a$, the next state $\vs'$ follows
\begin{eqnarray}
   p_\btheta(\vs'|\vs,a) 
&=  \prod_j p(s'_j|\vs,a) \\
p(s'_j|\vs,a) &= \frac{1}{Z_j}
\prod_{i: j(i) = j} p_i(s'_j|\vs,a)^{\theta_i}.
    \label{eq:poe-world}
\end{eqnarray}
where $j(i)$ is the target feature dimension
modeled by expert $i$.
Learning with PoE-World means generating experts $p_i$ with LLMs and estimating weights $\btheta$, which requires little data (few $(\vs,a,\vs')$ triples) because it is not learning a fully parametric model.
The model assumes the state features are conditionally independent;
this  makes it tractable
 to compute the partition function, $Z_j$,
 and hence we can perform
 maximum likelihood estimation (MLE) of the weights through gradient descent.
On object-centric Atari (OCAtari) \citep{delfosse2023ocatari}, PoE-World takes only a few minutes of gameplay to assemble a working world model.
We use PoE-World to learn an abstract world model.


\section{Method}



We propose AgentOWL (\textbf{O}ption and \textbf{W}orld model \textbf{L}earning Agent), an agent that sample-efficiently learns a sequence of options $\left\{ o_i \right\}$, given a goal-conditioned MDP and a sequence of goals $\left\{ g_i \right\}$. We describe our abstract world modeling approach embedded in AgentOWL in \cref{sec:awm} and then the full AgentOWL in \cref{sec:option_learning}.

\subsection{Abstract World Modeling}\label{sec:awm}

\input{figs/algorithm_pseudocode}

How should we abstract the state in order to reason about the effects of options?
Each goal predicate must be in the state abstraction in order to successfully capture how each option transforms the state.
In principle, further predicates may be important to include so that the abstract state is sufficiently informative, but recall that the abstract world model conditions on the current state $\vs$, so any further features can still be extracted from the current state.
We therefore define a state abstraction using just the goal predicates:
\begin{align}
\vf(\vs)&=\left( g_1(\vs), g_2(\vs), g_3(\vs), \ldots \right)
\end{align}
Next, we need an abstract world model that can be used for model-based lookahead.
We learn $p_o(\bdf'|\bds)$ using PoE-World \citep{piriyakulkij2025poeworld}, because by using symbolic programs to represent the abstract dynamics of the world, we can generalize more strongly from fewer examples.
Indeed, symbolic rules have long been an attractive representation for modeling coarse-grained world dynamics~\citep{fikes1971strips, mcdermott20001998}.

But even using symbolic programs, abstract states contain many abstract features, requiring many samples to learn. 
To maintain sample efficiency,
we impose a  ``frame axiom prior'' on the abstract world model, which biases it toward believing that option $o_i$ tends to change $f_i$ (from achieving $g_i$), but does not usually change $f_j$ for $j\not=i$.
The frame axiom prior is implemented by incorporating $p(\theta_i)$ into \cref{eq:poe-world}, turning weight optimization into a maximum a posteriori estimation (MAP) instead of MLE.
We use Gaussian priors $p(\theta_i) = \mathcal{N}(\mu, \sigma^2)$ with $\sigma=0.1$ and $\mu=0.5$ for experts that do not change $f_j$, and $\mu=0.001$ for the ones that do.
This ``frame prior'' is commonly used in the planning community, as it is employed, in a much stronger form, in PDDL \citep{mcdermott20001998}.

PoE-World yields  $p_o(\vf'\mid\vs)$, but  learning only this conditional distribution is insufficient because it cannot chain together several options:
After running the first option in state $\vs$, we arrive in $\vf'$, but predicting the effect of a second option might need the full state $\vs'$.
We heuristically predict $\vs'$ from $\vf'$ using a kernel density estimator $w(\vs'\mid\vf')$ that samples full states $\vs'$ given an abstract state $\vf'$:
\begin{align}
p_o(\vs'\mid\vs)\approx \mathbb{E}_{\vf'\sim p_o(\cdot\mid\vs)}\left[ w(\vs'\mid\vf') \right]
\end{align}
Note this is approximate: $\vs'$ generally depends on $\vs$ and $o$, even conditional on $\vf'$.
This approximation  is common in the hierarchical decision-making literature, where $w$ is called a \emph{weighting function}~\citep{bertsekas1995, li2006towards}. 
Weighting functions allow sampling states from an abstract state without learning the MDP transition function, and without training a parametric generative model over the raw state space, both of which would require enormous data.
To the extent that the environment can be accurately modeled using $p_o(\vf'\mid\vs)$ where $p_o$ only depends on the state abstraction $\vf(\vs)$, this approximation becomes exact.
\cref{app:awm} contains implementation details of our abstract world modeling approach.

\subsection{Joint Learning of Hierarchical Neural Options and Abstract World Model}\label{sec:option_learning}

Using this world modeling setup,
we now introduce the full AgentOWL.
It iteratively trains the next option to achieve the next goal (and in its world model) by calling \cref{fig:pseudocode}, whose three main ideas are described below.

\paragraph{Model-based exploration.}
Intuitively, planning in our world model should offer good guidance to a model-free policy;
we can weigh trajectories in imagination before deciding what to try in the real world.
Concretely, we run RL 
(specifically, deep Q-learning (DQN))
in the abstract world model
yielding a policy $\pi^{wm}$ (\cref{fig:pseudocode} line 15).
Note that this is computationally cheap,
since the
abstract world model $T$ takes large steps,
and is defined on a fairly low-dimensional symbolic state space,
which allows us to use simple MLPs
to represent the policy.\footnote{
Note $\pi^{wm}$ could be computed using a different strategy, such as planning in the world model, rather than RL in the world model. We leave exploring these alternatives for $\pi^{wm}$ for future work.
}

The resulting policy, $\pi^{wm}$,  
serves as an exploration policy for training a policy in the real world, $\pi^{real}$, that learns to achieve a goal.
More precisely, each option comprises a policy and goal, $o = (\pi, g)$, and we further decompose the policy into $\pi=(\pi^{real}, \pi^{wm}, \epsilon)$, where $\epsilon$ is the probability of taking exploratory actions (actions the world model predicts):
\begin{equation}
    \pi(a\mid \vs) = (1-\epsilon)\;\pi^{real}(a\mid \vs)+\epsilon\; \pi^{wm}(a\mid \vs)
\end{equation}
The decomposition ensures we can still learn a good policy even with imperfect world model. 
By annealing $\epsilon$ from $1$ to $0$, the agent eventually stops relying on $\pi^{wm}$ and falls back on fully model-free RL learning of $\pi^{real}$.
The reason we do this is that
model-based learning is known to be sensitive to model inaccuracy
\citep{gu2016continuous,janner2019trust}.

We note that for AgentOWL, each policy has its own set of weights; 
there is no weight sharing between the policies. 

\paragraph{Hypothesizing sub-options to achieve a target goal.}

Planning (or RL) to achieve a new goal is challenging unless we \emph{already} have an option which reaches that goal, which therefore could serve as a sub-option.
Absent such sub-options, the agent would need to reason about how its low level actions could be used to reach the new goal, defeating the whole point of a temporally abstract world model.
For example, if we have a sub-option to ``pick up the cup'' and a target goal of ``fill the cup with water'', a successful plan might first ``pick up the cup''  followed by a long sequence of low level actions.

To shorten our abstract plans, and help ``plan in the now'' \citep{kaelbling2011hierarchical}, we let the agent hypothesize new sub-options that aim to achieve the target goal given that certain preconditions are satisfied (see \cref{fig:learning} (right) and \cref{fig:pseudocode} line 10-13).
We use LLMs to propose the preconditions, $h$, of a new option $o_{h\rightarrow g}$. This new option, and its corresponding hypothetical option model $p_{o_{h\rightarrow g}}$, is then added to the set of options $\Omega$ and the abstract world model $T$ respectively.

For this work, we restrict preconditions to the form $h(\vs) =\vf(\vs)_i$, representing the completion of the sub-goal with index $i$. Concretely, we prompt an LLM to pick, among the sub-goals that the agent can already achieve with existing sub-options $\Omega$, a sub-goal that would be useful towards the target goal.
In the prompt, we include a sampled state from our set of seen transitions $D_{\Omega}$ to be included as part of the prompt. 
If the game has multiple rooms, we sample one state for each room (each state contains a ``room number'' object, so this can be easily done).
The exact prompt used can be found in \cref{app:prompt}.

\paragraph{Stable training of hierarchical options.}

Hierarchical option training is done in \cref{fig:pseudocode} line 18 using a hierarchical version of DQN. 
It proceeds similarly to typical DQN: executing the policy to collect data in the replay buffer and optimizing the policy using samples from the replay buffer.
However, in hierarchical DQN, the execution is hierarchical (\cref{fig:overview} right); the agent executes the root-level option, $o_g$, which then recursively calls sub-options until a primitive action is executed. 
We also assign each option its own replay buffer to keep the data it collects with its own policy.\footnote{Note that we could be more sample-efficient if we maintain a single shared replay buffer. We leave this for future work.}
Each time an option has collected enough new datapoints, the agent optimizes the option's policy weights for a fixed number of steps. 
Because of hierarchical execution, any sub-option may collect data and have its weights updated.

Nevertheless, hierarchical DQN can be unstable, because each higher-level option faces a non-stationary environment: Training lower level options changes the transition dynamics as seen by higher level options
~\citep{nachum2018data}. 
To mitigate this instability, note that an option's policy stabilizes once it has been trained with enough samples, or it reliably achieves its goals. 
We therefore disregard episode data for option training which contains an execution of at least one sub-option $o$ with $n_o < n_{threshold}$ and $\delta_o < \delta_{threshold}$, where $n_o$ is the number of samples the option has been trained with, $\delta_o$ is the option's goal completion rate over the 100 most recent episodes, and $n_{threshold}, \delta_{threshold}$ are hyperparameters. 
More details on stable Hierarchical DQN can be found in \cref{app:dqn}. 
\input{figs/block_stacking}
\input{figs/hard_goals}

\section{Experimental Results}

\paragraph{Experimental setup.} 
Each agent will be given an ordered sequence of target goals $(g_1, ..., g_n)$. 
The task for each agent is to ``master'' a set of neural options that correspond to the target goals, $(o_1, ..., o_n)$. 
We consider an option ``mastered'' when the goal completion rate of that option surpasses a threshold $\delta_{threshold}$.
For practicality, we approximate the goal completion rate by averaging goal completions over the 100 most recent execution of the option and set $\delta_{threshold} = 0.5$.
We evaluate the number of environment steps each agent uses to master this set of neural options.  


We conduct our experiments on a subset of object-centric Atari games and a simulated robotic block stacking environment.
In both domains, a state is a set of objects with primitive features, and goals are given as a sequence ordered by difficulty (for reasons discussed in \cref{sec:limitation}), so that earlier goals serve as stepping stones for later ones.

\textbf{Domain: Object-centric Atari (OCAtari)} \citep{delfosse2023ocatari} provides an object parser on top of Atari games \citep{bellemare13atari}, transforming the inputs from pixels to sets of objects.
Each object is described by object type (player, platform, ladder, etc.) and bounding box coordinates; we treat these values as primitive features.

We evaluate on the subset of games commonly used to study hard exploration in RL \citep{aytar2018playing, ecoffet2021first, hosu2016playing}, specifically Montezuma's Revenge, Pitfall and Private Eye, selected based on available computational resources.

For each game, we construct a small sequence of goals with the following procedure:
We manually select a few rooms in each game, as each may have many rooms, e.g., Pitfall contains 255 rooms.
We then define our list of goals as touching each possible object within these selected rooms. 
Finally, we manually order these goals by difficulty.

\textbf{Domain: Simulated robotic block stacking (\Cref{fig:block_stacking})} is a simulated tabletop manipulation domain built in PyBullet \citep{coumans2016pybullet}, featuring a robotic arm with a gripper and three cubes (red, green, and blue) on a table.
Similar to OCAtari, a state is a set of objects, namely the end effector and the three blocks, each described by its position, discretized to integer, together with a gripper open/closed feature.
The agent has nine discrete actions: a no-op, 6 movement actions along $\pm x$, $\pm y$, and $\pm z$, and two gripper actions (open/close).

The goal sequence consists of $15$ goals in three tiers: lifting a block ($3$ goals), stacking one block on another ($6$ ordered pairs), and building a three-block tower ($6$ orderings).
Unlike the Atari goal sequences, this goal sequence was produced by an LLM prompted with a description of the domain.
All methods attempt the goals in this order, but move a goal to the back of the queue if it fails to master the goal within a training budget.

Full details on the experimental setup and domains can be found in \Cref{app:exp} and \Cref{app:block_stacking}.




\input{figs/main_results}

\paragraph{Baselines.} 
\textbf{Rainbow DQN} \citep{hessel2018rainbow} is an improved version of DQN (Deep Q-Network) \citep{mnih2015human}, a standard off-policy RL algorithm commonly used in discrete action settings. 
All DQNs used in the paper are Rainbow DQNs.
\textbf{Goal-conditioned DQN} is DQN with weight sharing between the policies of the options. 
Specifically, instead of learning $\pi_1(a\mid\vs), ..., \pi_n(a\mid\vs)$, we seek to learn a goal-conditioned policy $\pi(a\mid\vs,g)$.
\textbf{Hierarchical DQN} is DQN whose policy has an action space that includes previously learned sub-options. 
Hierarchical DQN can be seen as AgentOWL without its abstract world model. 
Implementation details of DQN and Hierarchical DQN can be found in \cref{app:dqn}.

For the main results, we use Gemini 2.5 Flash as a sub-goal proposer for AgentOWL, but we also rerun AgentOWL with Qwen3.5-9B to show that the method does not require a strong LLM in \Cref{app:cheap_llm}.

\input{figs/ablation_results}

\paragraph{Skill acquisition results.}

As shown in \Cref{fig:main_results}, AgentOWL masters the highest number of options for most environment steps.
Although the baselines without hierarchical options seem to be better than AgentOWL at lower training sample budget for some environments, their performance plateaus at a much lower number of options mastered compared to that of AgentOWL.
Qualitatively, these baselines fail to acquire options for harder goals (\cref{fig:hard_goals}) because they never discover any action sequence that can achieve the harder goals, as the number of possible action sequences grows exponentially with the number of steps needed to achieve the goals.
AgentOWL, on the other hand, manages to accomplish a goal that requires a long, complicated sequence of primitive actions by planning abstractly with higher-level options and executing the options hierarchically.
Abstract plans can be very short, allowing them to be found very easily in both the real world and the abstract world model.

We perform an ablation study in \cref{fig:ablate_results}.
Removing techniques introduced in \cref{sec:option_learning}---specifically LLM-based sub-goal proposal and hierarchial training stabilization---degrades AgentOWL's performance across most environment steps.
These ablated systems need more data to master the same number of options and in many cases, plateau at fewer mastered options than the full system---with one exception where removing stabilization in Private Eye does not significantly change results.
We also show that AgentOWL still outperforms the baselines when the order of goals is randomized in \cref{app:random_goal_order}.

\input{figs/implicit_learning}

\paragraph{Implicit, hierarchical learning of sub-options.} 

AgentOWL improves its sub-options through hierarchical DQN even when rewarded only for goals that do not correspond to them.
In \cref{fig:implicit_learning}, we perform an experiment to clearly demonstrate this implicit learning. 
We take an AgentOWL agent trained to achieve a goal sequence in Montezuma's Revenge (\Cref{fig:learning} left), keep its learned abstract world model, but re-initialize the policy networks of all options with random weights.
Then, we solely train this agent to master goal ``key'', i.e., to touch the key.
\cref{fig:implicit_learning} (right) shows the performances of many re-initialized sub-options increase significantly after AgentOWL is trained to achieve goal ``key''.
We observe that sub-options helpful to the target goal improve, while irrelevant options corresponding to sub-goals outside the successful trajectory (\cref{fig:implicit_learning} left) do not. 

Intuitively, the agent leverages the learned world model to help decide a sequence of sub-goals to pursue to eventually achieve the target goal.
This sequence of goals acts as a curriculum for the agent to follow.
The agent refines these options to help it eventually achieve the target goal.

\input{figs/zero_shot_generalization}

\paragraph{Zero-shot generalization to novel situations.}

Another benefit of having a world model is zero-shot generalization to novel situations.
Looking forward to more complex domains, we want agents that can flexibly complete goals in novel situations, such as picking up a cup in a novel kitchen, or clearing a randomly generated level in Minecraft or Nethack \citep{johnson2016malmo, kuttler2020nethack}.

Thus, we design two experiments\footnote{We only ran Private Eye and block stacking experiments due to time constraints, but we believe the results for Montezuma's Revenge and Pitfall would be qualitatively similar.}
to showcase AgentOWL's zero-shot generalization capabilities. 
First, we test zero-shot generalization to a novel starting state in Private Eye (\Cref{fig:zero_shot}). 
In OCAtari, the starting state of each game is always exactly the same. 
We show that after learning an option to travel from a new 
starting state back to the game's original starting state, AgentOWL can compose existing options to achieve target goals zero-shot without any additional training data. 
Second, we test zero-shot compositional generalization to an unseen environment layout in the block stacking domain (\Cref{tab:zero_shot_comp}).
Each agent has to stack the red block on the green block.
The red and green blocks each start at one of four positions, giving 16 starting layouts, and we hold out one layout from training.
AgentOWL learns an option that lifts the red block from every red position and another option that places it on the green block at every green position.
Composing the two, AgentOWL succeeds in all 100 evaluation episodes on the held-out layout.

\section{Related Work}

\paragraph{Abstract and symbolic world models.}

The advent of LLMs has sparked interest in symbolic world models. 
WorldCoder and GIF-MCTS \citep{tang2024worldcoder, dainese2024codeworldmodel} directly use LLMs to generate and refine a world model as a program. 
POMDP Coder 
\citep{curtis2024partially}
and CWM \citep{lehrach2025code} tackle the problem of partial observability. 
PoE-World
\citep{piriyakulkij2025poeworld}
and OneLife \citep{khan2025one} 
scale up symbolic world modeling with a product-of-experts world representation.

Like our work, several prior works have explored abstract world modeling with symbolic world representations.
DECKARD \citep{nottingham2023embodied} learns a directed acyclic graph as its abstract world model.
Ada \citep{wong2024ada} synthesizes PDDL as its abstract world model.
AgentOWL, on the other hand, learns a stochastic, symbolic world model capable of making both abstract and low-level predictions, offering greater representational power.

\nocite{ahmed2025synthesizing}

\paragraph{Option discovery and hierarchical RL.}

Option critic methods \citep{bacon2017option, harutyunyan2019termination, tiwari2019natural} learn both policies and termination functions of options end-to-end through gradient descent. 
Skill-chaining \citep{konidaris2009skill, bagaria2019option, bagaria2021robustly} focuses on discovering chainable options, where the termination set of an option is the initiation set of another.
Other hierarchical RL methods tend to have high-level policies and low-level policies, where the high-level ones either set goals or rewards for the low-level ones \citep{dayan1992feudal,kulkarni2016hierarchical,vezhnevets2017feudal, li2019hierarchical, hafner2022deep} or directly select which low-level ones to execute \citep{florensa2017stochastic,heess2016learning,eysenbach2018diversity}.
Our work takes inspiration from these prior works but differs in how we learn deeply hierarchical options, as opposed to the common two-level hierarchy of policies. 

\nocite{nachum2018data}

\paragraph{LLMs for RL.} 

Existing works have explored the use of LLMs in RL agents in many ways, including assisting with reward design \citep{kwon2023reward,ma2023eureka,klissarovdoro2023motif,castanyer2025arm}, serving as policies or policy generators \citep{yao2022react,wang2023voyager,liang2022code}, and producing high-level plans \citep{ahn2022can, huang2022language, singh2022progprompt, song2023llm}.
AgentOWL leverages LLMs for world modeling and sub-goal proposal, but can also benefit from LLM-based reward design if reward functions are not provided.
Importantly, AgentOWL learns neural policies through environmental interaction rather than relying on LLMs to directly output action sequences or policies. 

\section{Limitations and Future Direction}\label{sec:limitation}


While AgentOWL efficiently learns abstract
world models and hierarchical neural options in our setting, there are limitations. 
First, we assume the given sequence of goals is ordered by difficulty.
What made AgentOWL effective is the ability to use sub-options to help achieve a hard goal.
Consequently, AgentOWL can fail to achieve a challenging goal without first learning to accomplish its prerequisite sub-goals.
In the future, we hope to use ideas
from curriculum learning \citep{bengio2009curriculum}
to automate this.


Second, we assume the number of goals is relatively small ($< 100$).
While AgentOWL's model-based exploration keeps environment interactions from scaling linearly with the number of goals, training compute still does.
Thus, currently, we  reduce the number of environment samples at the expense of increasing compute, which is the right tradeoff only when compute is cheaper than data.
Incorporating option affordances \citep{khetarpal2020can, khetarpal2021temporally} to reduce the number of applicable options could be a fruitful direction.

Lastly, we assume symbolic input to the abstract world model, using OCAtari instead of pixel-level Atari.
Symbolic input permits  learning world models like PoE-World, which is much more sample-efficient than learning a pixel-level world model.
There is ongoing effort to learn abstract symbolic world models directly from pixels \citep{liang2024visualpredicator, liang2025exopredicator,athalye2024pixels}, but merging that line of work with option training remains open.
Purely neural world models have made great strides~\citep{ball2025genie,alonso2024diamond}, but do not learn explicit symbolic abstractions that can be used to reason over long  horizons. 
Leveraging these purely neural models to efficiently learn neurosymbolic world models offers another path forward.

Limitations aside, AgentOWL's successful results emphasize that option and world model learning are deeply intertwined, and that skills are fundamental to an agent's understanding of its environment.
We hope this insight draws attention to several underexplored research problems. 
For example, if we allow goals and corresponding skills to grow over time, our abstract model faces an ever-changing state and action space.
How do we perform efficient world modeling in this online learning setting? 
With large sets of abstract features and skills, how do we reason over only relevant ones to save computation? 
Answering these questions could make systems like AgentOWL much more powerful.

\section*{Acknowledgments}

Kevin Ellis and Wasu Top Piriyakulkij acknowledge support from an NSF CAREER grant.

\nocite{parr1998hierarchical}
\nocite{dietterich2000hierarchical}
\nocite{kulkarni2016hierarchical}
\nocite{bagaria2021skill}
\nocite{eysenbach2018diversity}

\bibliographystyle{unsrt}
{\small \bibliography{example_paper}}

\clearpage

\appendix

\section{Technical appendices}
\subsection{DQN and Hierarchical DQN}\label{app:dqn}

Here we describe our implementation of Rainbow DQN \citep{hessel2018rainbow} which is used in all baselines, including AgentOWL itself. 

\paragraph{Neural network architecture.} 
Rainbow DQN uses dueling network architecture \citep{wang2016dueling}
We use a 2-layer MLP as our feature extractor module. 
The hidden feature size is 256 for $\pi^{real}$ and 128 for $\pi^{wm}$ for both layers (so always 256 for baseline methods).
Each linear layer is followed by a layer normalization (LN) layer \citep{ba2016layer}, as recent studies found incorporating LN in the network of a deep RL algorithm to be highly beneficial \citep{lyle2023understanding, lyle2024disentangling}, and then a ReLU layer.
After the feature extractor, the value and advantage layer is also 2-layer MLP with hidden feature size half of what is used in the feature extractor, but the linear layers are replaced with noisy linear layers \citep{fortunato2018noisy}, and LN and ReLU are only applied the first layer, since the last layer produces output.

For goal-conditioned DQN, we also learn an embedding for each goal and concatenate it to the input to the network.

\paragraph{Training.}

We base our implementation of (Rainbow) DQN and hierarchical DQN off of Stable Baseline's implementation \citep{stable-baselines3}.
The pseudocode for DQN is and hierarchical DQN is at \Cref{fig:dqn_pseudocode} and \cref{fig:hdqn_pseudocode} respectively, with \cref{fig:helper_pseudocode} and \cref{fig:helper_pseudocode_2} describing the helper functions used in hierarchical DQN. 
The common hyperparameter for all variants of DQN is listed at \cref{tab:hyperparameters}.
There are extra hyperparameters for hierarchical DQN \cref{tab:hyperparameters_hdqn} and overridden hyperparameters for training $\pi^{wm}_g$ \cref{tab:hyperparameters_wm}.

\paragraph{Goal heuristics.} Additionally, to speed up training, we have a heuristic for each environment, described in detail in later sections.
We incorporate this heuristic $h$ in our Q network as follows: 
\begin{align}
    Q^{new}_\psi(s, a) = Q_\psi(s,a) + h_g(s)
\end{align}

Intuitively, a freshly initialized neural network typically outputs values around $0$. 
Adding $h_g(s)$ to it makes $Q_{new}$ outputs values around $h_g(s)$ instead.
Over time, as we train $Q^{new}$ more and more, the neural network learns to correct the heuristics value so that the output is close the true Q value \citep{ng1999policy}.

\subsection{Experimental setting and OCAtari details}\label{app:exp}

Note that the Arcade Learning
Environment (ALE) \citep{machado18arcade} code uses GPL-2.0 license, and OCAtari code \citep{delfosse2023ocatari} uses MIT license.

\paragraph{Goal sequences for each game.}

For each game, we select a few rooms and define goals as touching objects within them. 
Then, we manually order the goals based on difficulty.
Additionally, we give short natural language names to each goal, such as `top\_left\_plat', `mid\_right\_wall', etc.
We display the screenshots of the rooms with goal objects labeled with their order in the goal sequence at \cref{fig:mr_goals}, \cref{fig:pitfall_goals}, and \cref{fig:privateeye_goals}.
We note that the orders within each room in Pitfall and Private Eye tend not to matter too much as they are roughly the same in difficulty level.

\paragraph{Goal heuristics.} For each game, we use a heuristic based on the Manhattan distance between the player object and the goal object. Specifically, we let the heuristic function $h$ be $h(s) = 5 \cdot (1-\frac{ManhattanDist(s_{player}, s_{g-object)}}{400})$. 
In OCAtari, Manhattan distance between any two objects never exceeds 400. 
If the goal object is not visible in $s$, e.g., the goal object is in another room in the game, we let the heuristics value equal to 0.

\paragraph{Vectorizing list of objects.}

We vectorize each input object list into a fixed-size feature vector for the DQN policy. 
By using the maximum count for each object type, we assign each object to a unique location in the observation space. 
For example, with one player and up to two enemies, where each object is vectorized into a 8-dimensional feature vector, the observation space has size 24: the player occupies indices 0-8, the first enemy occupies indices 9-16, and so on.

To vectorize each object, we take its x and y coordinates and encode each value into a 4-dimensional vector using a positional encoder implemented with sine and cosine functions, commonly used in LLMs.

Additionally, we include the goal values, i.e., the abstract state $\vf(\vs)$, in the feature vector as well, so the feature vector contains both low-level and abstract features. To better match the dimensionality of the abstract features to that of the low-level features, we duplicate each abstract feature by 4.

\paragraph{Observation space optimization.}

To optimize running time, we choose to disregard static (non-moving) object types, such as platforms, ladders, etc., from the vector discussed above, as all states have the same information for these objects. This speeds up our code significantly, reducing the low-level feature vector size from $248, 344, 408$ to $24, 176, 328$ for Montezuma's Revenge, Pitfall, and Private Eye, respectively.

\paragraph{Episode timeouts.}

Since stable hierarchical DQN only adds data to the replay buffer once an episode ends, we need to make sure that episodes do not go on forever. Toward that end, we set a maximum time limit for all episodes in a (real) environment to be $1,000$ environment steps. On the other hand, when we treat our abstract world model as a simulated, abstract environment, we set the time limit to be only 4 environment steps, as this simulated environment is very abstract and only requires very short action sequences to achieve the goal.

\subsection{Simulated robotic block stacking experiments details}\label{app:block_stacking}

\paragraph{Environment.}
The blocks are $5$\,cm cubes.
The robot arm is allowed to move within a $60$\,cm cube workspace in front of the robot base, with a table at the bottom.
Each movement action moves the end-effector by $5$\,cm along one axis.

\paragraph{Goal heuristics.}
The heuristic is based on distances between the end effector and the relevant blocks and between the blocks themselves, and is scaled to lie in $[0, 0.5]$.

\paragraph{Vectorizing list of objects}
We vectorize states as in OCAtari (\cref{app:exp}), with one change to the positional encoder.
With the encoder used for OCAtari, whose coordinates range over hundreds of pixels, some pairs of positions that are far apart in this $60$\,cm workspace receive more similar encodings than pairs that are close together.
We therefore rescale coordinates by $0.1$ and use base $5$ instead of $10{,}000$ in the sine and cosine functions, which makes the distance between encodings increase with the distance between positions.

\paragraph{Round-robin goal scheduling with exponential give-up mechanism.}
All methods in the main block stacking experiment (\cref{fig:main_results}) use an exponential give-up mechanism, with the goals queued in the order generated by the LLM.
Each attempt at a goal gets a budget of $250$K environment steps times a multiplier that starts at $1$.
If the goal is not mastered within the budget, it moves to the back of the queue.
AgentOWL, hierarchical DQN, and goal-conditioned DQN then revert to a snapshot of the agent taken just before the attempt, so a failed attempt cannot damage previously learned options.
DQN, which trains an independent policy for each goal, instead restarts the goal from scratch with a new random seed on its next attempt.
The multiplier doubles after a full pass over the queue in which no goal is mastered and resets to $1$ whenever a goal is mastered.
Because the budget keeps growing, the initial budget is not a fragile hyperparameter: it affects how quickly AgentOWL learns but not its asymptotic performance.

\paragraph{Option resets.}
Two safeguards keep option training from getting stuck.
If an option that has not yet been mastered goes $100$K steps without a single success, we re-initialize its network and empty its replay buffer.
If a mastered option fails $20$ episodes in a row, we restore that option from the most recent snapshot saved after a goal was mastered, the last state in which all options were known to work together.

\paragraph{Neural network architecture.}
The main block stacking experiment uses the same 2-layer MLP feature extractor as OCAtari (\cref{app:dqn}).
The zero-shot compositional generalization experiment (\cref{tab:zero_shot_comp}) replaces it with a transformer encoder for both AgentOWL and DQN.
Each object in each stacked frame becomes one token, and the abstract features become one more token.
Each token is linearly projected to the hidden size of the MLP, and a learned per-token embedding is added.
A learned classification token is prepended to the sequence, which then passes through a $2$-layer transformer encoder with $4$ attention heads and feed-forward size twice the hidden size.
The output at the classification token, after layer normalization, goes to the same dueling value and advantage layers as in the MLP version.
Pooling through a learned classification token, rather than averaging all tokens, lets the network ignore objects that are irrelevant to the current goal.

\paragraph{Zero-shot compositional generalization setup.}
In this experiment, the red block starts at one of four positions with $x \in \{20, 25\}$, $y \in \{15, 20\}$, and the green block also starts at one of four positions with $x \in \{20, 25\}$, $y \in \{40, 45\}$, while the blue block stays at its default position.
We hold out one layout, $x=25$, $y=20$ for red and $x=25$, $y=40$ for green, and let the methods train on the other 15.
AgentOWL learns the goal sequence $lift\_red$ and $red\_on\_green$, while DQN learns the goal $red\_on\_green$ directly.
Compared with the main experiment, we raise the mastery threshold to $\delta = 0.95$, shorten real-environment episodes to $50$ steps, lower the learning rate to $5 \times 10^{-5}$, enlarge the replay buffer to $40{,}000$ transitions, and update the networks every $256$ environment steps instead of $64$, taking $0.25$ instead of $1$ gradient step per environment step.

\subsection{Abstract world modeling implementation details}\label{app:awm}

\paragraph{Learning $p_o(\vf'|\vs)$ with PoE-World.}

We first assume the following structure for $p_o(\vf'|\vs)$:
\begin{align}
    p_o(\vf'|\vs) \triangleq p_o(f'_o|\vs) (\prod_{i \neq o} p_o(f'_i|f'_o, \vs))
\end{align}

Then, each $p_o(f'_o|\vs)$ and $p_o(f'_i|f'_o, \vs)$ can be modeled with a product of experts and learned with PoE-World. 
And as mentioned in the main text, for $p_o(f'_i|f'_o, \vs)$, we further incorporate Gaussian priors with $\sigma=0.1$ and $\mu=0.5$ for experts that do not change $f_i$, and $\mu=0.001$ for the rest.

The intuition behind the above structure is the model should first predict $f'_{o}$ which corresponds to the result of executing option $o$, whether or not it will achieve its corresponding goal $g$, with $p_o(f'_{o}|s)$.
Based on the result of the execution, the model can then predict the rest of the abstract features with $\prod_{i \neq i_o} p_o(f'_i|f'_o, s)$.

\paragraph{Expert generation for PoE-World.}

For $p_o(f'_{o}|s)$ of each option $o$, we prompt an LLM to synthesize a set of possible preconditions: Gemini 2.0 Flash-Lite with the prompts in \cref{prompt:poe_world_1,prompt:poe_world_2} for OCAtari, and Gemini 2.5 Flash with a block stacking version of these prompts for the block stacking domain.
We then add one expert per precondition that sets $f'_{o}=1$ if that precondition is true.
We also add a ``blanket'' expert that sets $f'_{o}=0$ without any precondition.
Both types of experts have Gaussian weight prior with $\mu=0.5$, but the blanket expert has $\sigma=0.001$, while the experts with preconditions have $\sigma=0.1$.
We give low $\sigma$ to the blanket expert because we want its weight stays close to $0.5$.
If we allow the weight for the blanket expert to change too much, it might become 0 when only positive examples are observed during fitting, causing the model to incorrectly conclude that option $o$ will always succeed without any preconditions.

For $p_o(f'_i|f'_o, s)$ of each option $o$, we generate three experts: $f'_i=f_i$ (no change), $f'_i=1$, and $f'_i=0$.
As mentioned in \Cref{sec:awm}, we give a Gaussian weight prior with $\mu=0.5,\sigma=0.1$ for the first type of expert (no change), and $\mu=0.5,\sigma=0.001$ for the latter two types.


\paragraph{Weighting function.}

We implement weighting functions straightforwardly as a lookup table with the keys being the abstract states and the values being the corresponding low-level states.
We only keep a single low-level state for each abstract state in the lookup table, so there is no sampling.

\paragraph{Undefined distribution and partial state.}

It is possible that in the weighting function, we try to sample from $w(\vs\mid \vf)$ when we have not yet seen a $\vs$ that corresponds to $\vf$, meaning our lookup table for the key $\vf$ is empty.
When that happens, we leave $\vs$ unspecified.
Concretely, we assign $None$ values to all primitive features. 
We call states with $None$ values for some features, partial states.
Arriving at partial states in the world-model-simulated environment does not terminate the episode right away.
The episode only terminates if the experts in $p_o(\vf'|\vs)$ try to access the features with $None$ values.

Additionally, we also set $f'_i=None$ for all $i\neq o$ when we sample from $p_o(\vf'| \vs)$ but get $f'_o = 0$.
We implement this mechanism because options can lead to highly unpredictable states when they are not successful in achieving their corresponding goals.

\subsection{Experiments with a small language model}\label{app:cheap_llm}

AgentOWL relies on an LLM for sub-goal proposal and world model learning, and our main Atari experiments use Gemini 2.5 Flash for the former and Gemini 2.0 Flash-Lite for the latter, which costs around \$1 in API calls per run.
To test whether AgentOWL needs a strong LLM, we rerun AgentOWL on Montezuma's Revenge with Qwen3.5-9B, a small open-weight model that fits on a single consumer GPU.
As shown in \cref{tab:cheap_llm}, AgentOWL with Qwen3.5-9B masters options for all 18 goals within 1M environment steps in every seed, matching AgentOWL with Gemini models, while every baseline plateaus at 13 goals.

\begin{table}[h]
\centering
\begin{tabular}{lc}
\toprule
Method & Options mastered after 1M steps \\
\midrule
DQN & $13.00 \pm 0.00$ \\
Goal-conditioned DQN& $13.00 \pm 0.00$ \\
Hierarchical DQN & $13.00 \pm 0.00$ \\
AgentOWL w/ Gemini models (\cref{fig:main_results}) & $\mathbf{18.00 \pm 0.00}$ \\
AgentOWL w/ Qwen3.5-9B & $\mathbf{18.00 \pm 0.00}$ \\
\bottomrule
\end{tabular}
\vspace{1em}
\caption{Number of options mastered after 1M environment steps in Montezuma's Revenge, mean $\pm$ standard error over 3 seeds. Replacing Gemini models (Gemini 2.0 Flash-Lite + Gemini 2.5 Flash) with the much smaller Qwen3.5-9B does not change AgentOWL's performance.}
\label{tab:cheap_llm}
\end{table}

\subsection{Random and LLM-generated goal order}\label{app:random_goal_order}

Our main experiments give AgentOWL a sequence of goals ordered by difficulty.
To relax this assumption, we run AgentOWL on Montezuma's Revenge with two other goal orders: a random order, and an order proposed by an LLM.
Since an agent might be asked to master a goal before it has the sub-options needed to make that goal tractable, we employ the same round-robin goal scheduling as \Cref{app:block_stacking}.

\cref{tab:random_goal_order} shows that AgentOWL still outperforms all baselines with either goal order, although it masters fewer goals than with the manual order under the same training budget.
In domains where the LLM has a strong prior, ordering the goals by difficulty with an LLM is also feasible: in our block stacking domain, the goals, their predicates, and their order are all generated by an LLM agent.

\begin{table}[h]
\centering
\begin{tabular}{lc}
\toprule
Method & Options mastered after 5M steps \\
\midrule
DQN & $13.00 \pm 0.00$ \\
Goal-conditioned DQN & $13.00 \pm 0.00$ \\
Hierarchical DQN & $13.00 \pm 0.00$ \\
AgentOWL w/ manually ordered goals (\cref{fig:main_results}) & $\mathbf{18.00 \pm 0.00}$ \\
AgentOWL w/ LLM-ordered goals & $16.33 \pm 1.67$ \\
AgentOWL w/ randomly ordered goals & $16.67 \pm 0.67$ \\
\bottomrule
\end{tabular}
\vspace{1em}
\caption{Number of options mastered after 5M environment steps in Montezuma's Revenge with different goal orders, mean $\pm$ standard error over 3 seeds. The baselines use the manual goal order.}
\label{tab:random_goal_order}
\end{table}

\subsection{Prompts for LLM-based sub-goal proposer}\label{app:prompt}

\cref{prompt:subgoal} contains the prompt used for LLM-based sub-goal proposer.

\subsection{Compute resources and execution time}

\paragraph{Compute Resources.}

Our experiments are run on 1 GPU (NVIDIA RTX A6000 or NVIDIA GeForce RTX 3090) with 4 CPUs (CascadeLake, IceLake, or SaphireRapids) and 48 GB memory. The method involves calling LLMs through OpenRouter API; each experiment costs around \$1 in API cost.

\paragraph{Execution time.}

Our method tends to take around 1 day for each experiment on Montezuma's Revenge and PrivateEye experiments and 2 days for each experiment on Pitfall.

\input{figs/dqn_pseudocode}
\input{figs/hdqn_pseudocode}
\input{figs/hdqn_pseudocode_2}
\input{figs/hdqn_pseudocode_3}
\clearpage
\input{figs/hyperparameters}
\input{figs/hyperparameters_hdqn}
\input{figs/hyperparameters_wm}
\clearpage
\input{figs/gen}
\input{figs/mr_goals}
\input{figs/pitfall_goals}
\input{figs/privateeye_goals}
\input{prompts/subgoal_proposer}
\input{prompts/poe_world_1}
\input{prompts/poe_world_2}

\clearpage
\input{checklist.tex}

\end{document}

%% file: math_commands.tex
\usepackage{amsmath,amsfonts,bm}

\def\eqref#1{equation~\ref{#1}}

\def\1{\bm{1}}

\def\vf{{\bm{f}}}

\def\vs{{\bm{s}}}

\DeclareMathAlphabet{\mathsfit}{\encodingdefault}{\sfdefault}{m}{sl}
\SetMathAlphabet{\mathsfit}{bold}{\encodingdefault}{\sfdefault}{bx}{n}

\newcommand{\bds}{\mathbf{s}}
\newcommand{\bdf}{\mathbf{f}}

\newcommand{\btheta}{{\boldsymbol{\theta}}}

%% file: abstract.tex
\begin{abstract}
Building agents that can perform new skills by composing existing skills is a long-standing goal of AI agent research.
Towards this end, we investigate how to efficiently acquire a sequence of skills, formalized as hierarchical neural options.
However, existing model-free hierarchical reinforcement algorithms  need a lot of data.
We propose a novel method, which we call
AgentOWL (\textbf{O}ption and \textbf{W}orld model \textbf{L}earning Agent),
that jointly learns---in a sample efficient way---an abstract world model  (abstracting across both states and time) and  a set of hierarchical neural options.
We show, on a subset of Object-Centric Atari games and a simulated robotic block stacking environment,
that our method can learn more skills using less data
than baseline methods and possesses learning and generalization capabilities that the baselines do not have.
\end{abstract}

\eat{
\begin{abstract}
Building agents that can perform new skills by composing existing skills is a long-standing goal of AI agent research. 
Towards this end, we investigate how to efficiently acquire a sequence of skills, formalized as hierarchical neural options.
Model-free reinforcement algorithms fail to learn deeply hierarchical options because the tradeoff where the number of training samples required scale linearly with number of sub-options available.
We propose a novel world modeling algorithm that is highly sample-efficient and embed it in an agent that jointly learns hierarchical options and an abstract world model, allowing us to perform model-based learning and bypass the mentioned tradeoff.
Our results on a subset of OCAtari games show that given a sequence of skills to learn, our agent acquire the highest number of skills under the same training sample budget with other baselines.
\end{abstract}}

%% file: figs/overview.tex
\begin{figure*}[t]
\centering
\includegraphics[width=\linewidth]{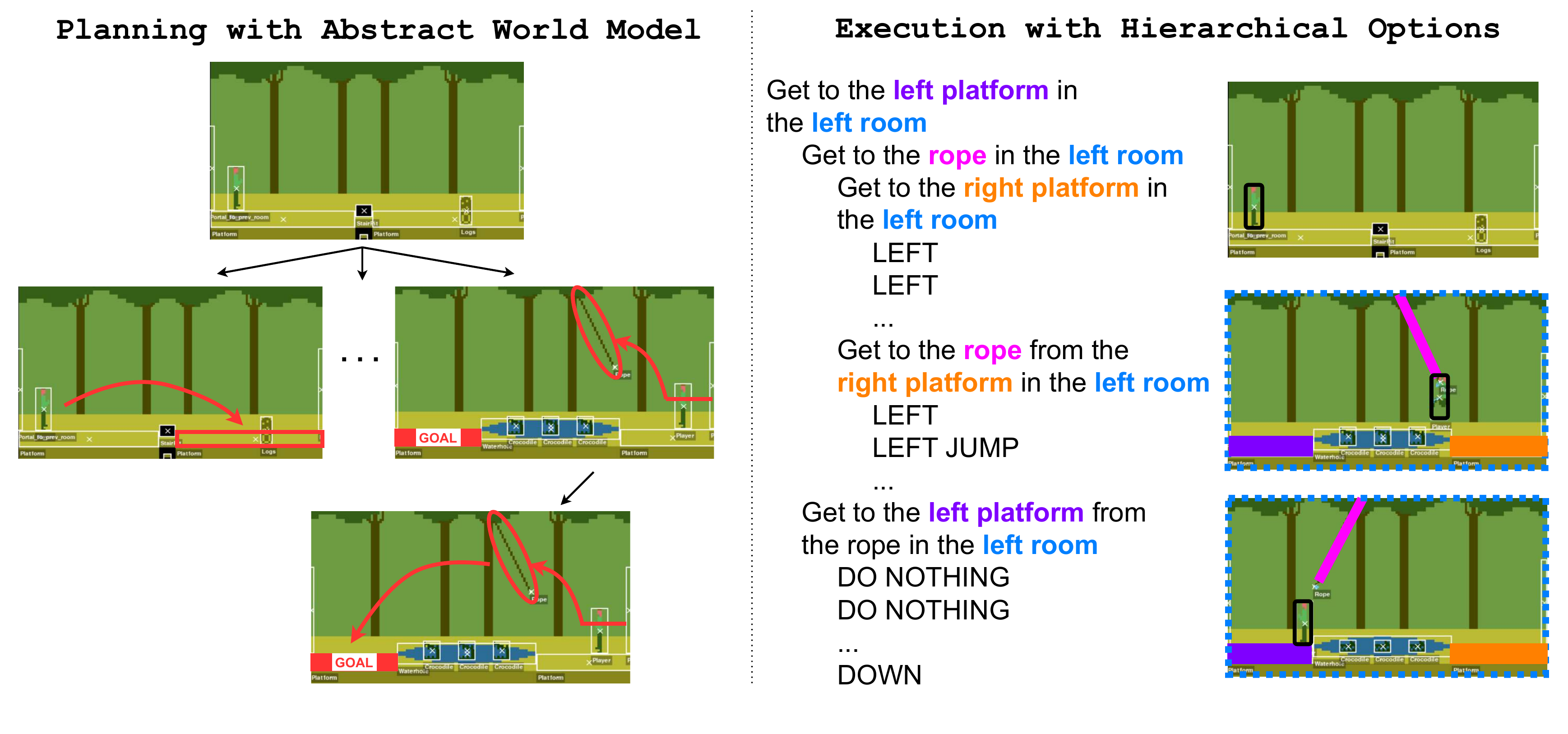}
\caption{Illustration of hierarchical planning and execution in AgentOWL where the goal is to go to the left platform in the left room.
Left: AgentOWL goes through possible plans and successfully makes a short plan (two high level steps) in its abstract world model to reach the goal.
Right: AgentOWL executes a hierarchical sequence of options. (The hierarchical structure is represented by the indentation.)
}
\eat{
\caption{Illustration of how AgentOWL
plans with its abstract world model (left) and executes its hierarchical options (right).
Left: AgentOWL plans at the abstract level with the abstract world model that knows the model of options. The abstract plan can be incredibly short (only two steps in the shown figure) as it abstracts the time in the actual environment.
Right: AgentOWL possesses deeply hierarchical options, as indicated by the indentation in the shown text. Following the abstract plan, AgentOWL execute options which calls sub-options which calls sub-sub-options and so on.
}
}
\label{fig:overview}
\end{figure*}

%% file: figs/learning.tex
\begin{figure*}[t]
\centering
\includegraphics[width=\linewidth]{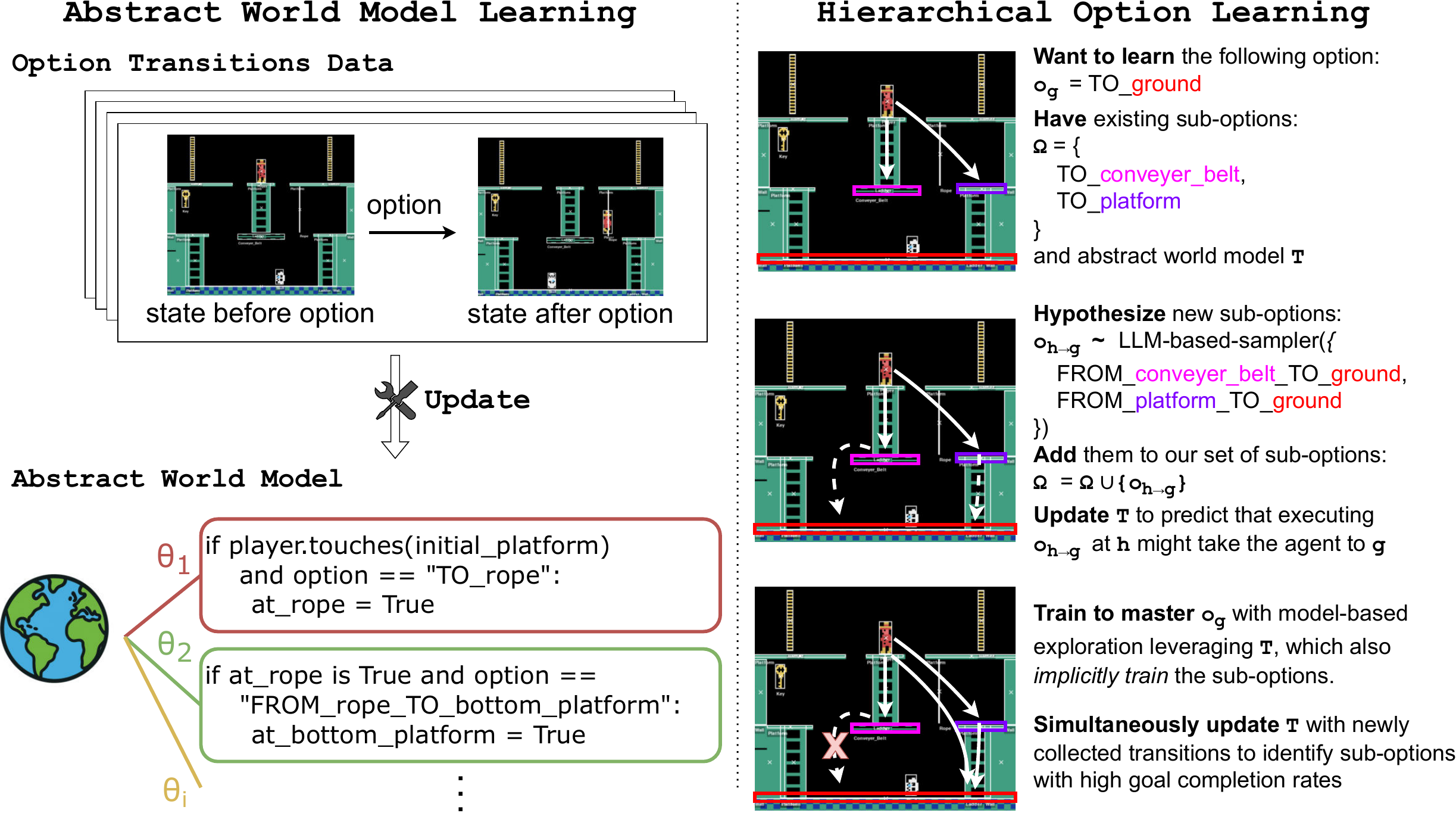}
\vspace{0.3em}
\caption{Illustration of how AgentOWL learns its abstract world model (left) and hierarchical options (right). 
Left: given a dataset of option transitions, we learn an abstract world model using 
(an extension of) 
the method
of \cite{piriyakulkij2025poeworld}.
Specifically, each expert is generated using LLM code synthesis, and the weight for each expert (denoted $\theta_i$) is learned
using gradient descent on the likelihood objective.
Right: AgentOWL learns a new option to achieve a new goal by leveraging previously acquired sub-options.
Specifically, given the goal,
we ask an LLM to hypothesize new sub-options (building on the already learned ones) that might help achieve the goal.
The agent trains to master the highest-level option, $o_g$, which implicitly trains these new sub-options, and simultaneously updates our abstract world model to identify sub-options with high goal completion rates.
Eventually, the agent masters the target option $o_g$ by composing good sub-options.
Detailed pseudocode can be found in \cref{fig:pseudocode}.}
\eat{
\caption{Illustration of how AgentOWL learns its abstract world model (left) and hierarchical options (right). 
Left: given a set of option transitions, we fit our abstract world model, partially expressed as a product of programmatic experts \cite{piriyakulkij2025poeworld}, to the data. 
Right: AgentOWL learns a new option to achieve a new goal by leveraging previously mastered sub-options. Given the sub-goals AgentOWL know it can achieve with the sub-options, it hypothesizes new sub-options that might be able to achieve the target goal given certain preconditions. AgentOWL learns whether each hypothetical sub-option is useful or not and uses the sub-options to learn the hierarchical option to achieve the target goal. Pseudocode for the learning algorithm can be founded at \cref{fig:pseudocode}.}
}
\label{fig:learning}
\end{figure*}

%% file: figs/algorithm_pseudocode.tex
\begin{algorithm*}[!t]
\caption{AgentOWL learning algorithm}
\small
\begin{algorithmic}[1]
\STATE Given a target goal to achieve $g$, a list of sub-options $\Omega$ and corresponding policies $\pi_\Omega= \{\pi_o\}_{o\in \Omega}$, a current abstract world model $T=\{p_o\}_{o\in \Omega}$, a set of seen option transitions $D_{\Omega}=\{D_o\}_{o\in \Omega}$, an environment $E$, and stabilization thresholds $n$ (number of samples) and $\delta$ (goal completion rate).
\STATE $\pi_g^{real} \gets InitPolicy(E, \Omega)$
\COMMENT{Initialize policy}
\STATE $\epsilon \gets 1$ 
\COMMENT{Initialize $\epsilon$ which will anneal over time to $0$}
\WHILE{$o_g$ for goal $g$ is not mastered}

\STATE $\Omega_g \gets \{o: g_o = g \mid o \in \Omega\}$ \COMMENT{Set of options whose corresponding goals match the target goal}
\STATE $\Omega_{stable-g} \gets \{o: n_o > n \text{ or } \delta_o > \delta \mid o \in \Omega_g\}$ 
\COMMENT{Subset of $\Omega_g$ with only ``stable'' sub-options}
\STATE $\Omega_{good-g} \gets \{o: \delta_o > \delta \mid o \in \Omega_g\}$ \COMMENT{Subset of $\Omega_g$ with only good, high-performing options}
\STATE \textit{\# If all relevant options are stable but none are good enough, then generate new sub-goal}
\IF{$|\Omega_g| = |\Omega_{stable-g}|$ and $|\Omega_{good-g}|= 0$} 
\STATE $h \gets$ $LLM(g, \Omega, D_{\Omega})$ \COMMENT{Sample a precondition with LLM}
\STATE $\Omega \gets \Omega \cup \{o_{h\rightarrow g}\}$ \COMMENT{Add a new sub-option $o_{h\rightarrow g}$ to $\Omega$}
\STATE $T \gets T \cup \{p_{o_{h\rightarrow g}}\}$ \COMMENT{Add the corresponding hypothetical option model $p_{o_{h\rightarrow g}}$ to $T$}
\STATE $\pi_g^{real} \gets InitPolicy(E, \Omega)$ \COMMENT{Reinitialize the policy as $\Omega$ changes}
\ENDIF
\STATE $\pi_g^{wm} \gets DQN(T, \Omega, R_g)$ \COMMENT{\parbox[t]{.60\linewidth}{Learn $\pi_g^{wm}$ from scratch in the abstract world model $T$ with $\Omega$.
}}
\STATE \textit{\# Train hierarchically to master an option for goal $g$ in $E$.}\\
\STATE \textit{\# This may change any $\pi \in \pi_\Omega$, in addition to $\pi_g$.}\\
\STATE $((\pi^{real}_g, \pi^{wm}_g, \epsilon), \pi_\Omega, D_{\Omega}) \gets HierarchicalDQN(E, \Omega, D_{\Omega}, \pi_\Omega, g,n,\delta, \pi_g=(\pi^{real}_g, \pi^{wm}_g,\epsilon))$ 

\STATE $T \gets PoEWorld(T, D_{\Omega})$ \COMMENT{Fit $T$ with the updated $D_{\Omega}$ using PoE-World}
\ENDWHILE
\STATE $o_g \gets ((\pi^{real}_g, \pi^{wm}_g, \epsilon), g)$
\STATE \textbf{return} $o_g, \Omega, T, D_{\Omega}$ \COMMENT{Return the target option $o_g$ and the updated $\Omega, T, D_{\Omega}$}
\end{algorithmic}
\label{fig:pseudocode}
\end{algorithm*}

%% file: figs/block_stacking.tex
\begin{figure*}[t]
\centering
\includegraphics[width=0.95\linewidth]{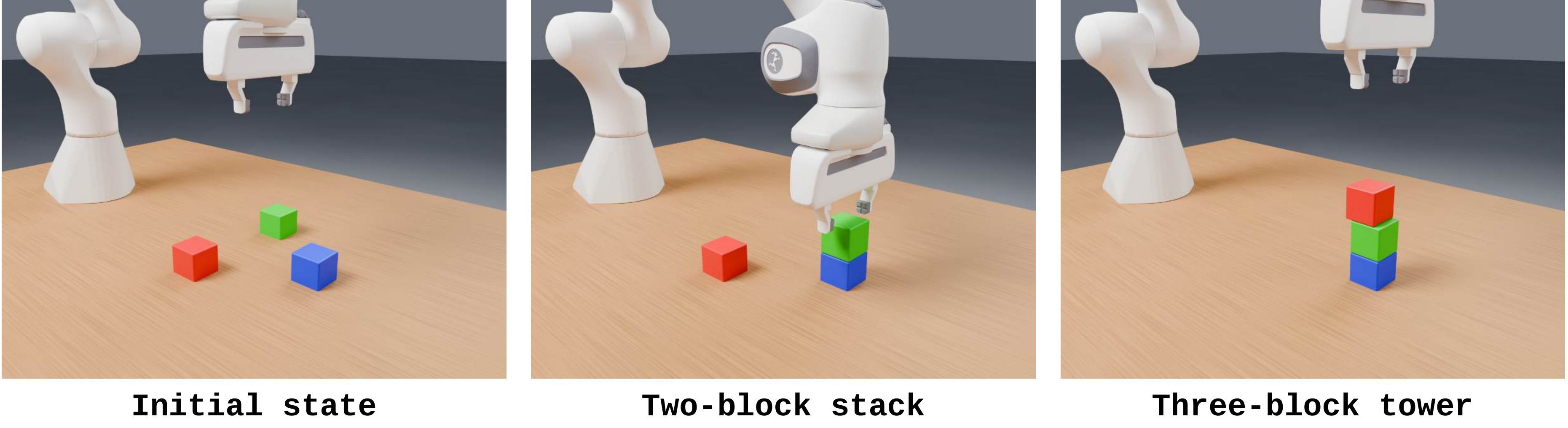}
\caption{Screenshots from the simulated robotic block stacking domain, showing the initial state and states that achieve a two-block and a three-block stacking goal.}
\label{fig:block_stacking}
\end{figure*}

%% file: figs/hard_goals.tex
\begin{figure*}[t]
\centering
\includegraphics[width=\linewidth]{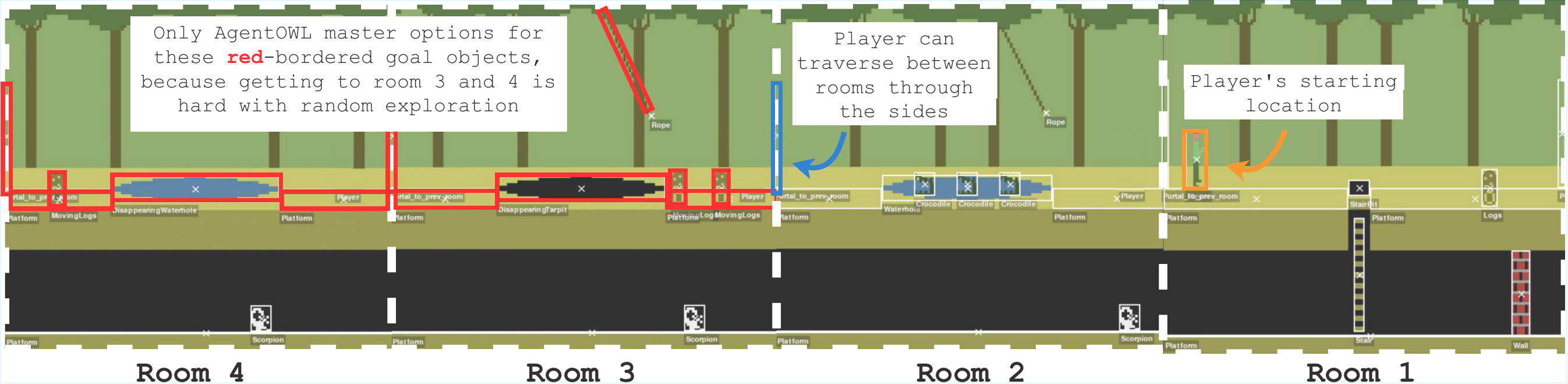}
\caption{Screenshots of 4 rooms of Pitfall stitched together. Player starts in Room 1 (rightmost) and can traverse to other rooms through the sides of the screen. Goals that only AgentOWL masters within 5M environment steps (\cref{fig:main_results} middle) are shown in red borders.} 
\label{fig:hard_goals}
\end{figure*}

%% file: figs/main_results.tex
\begin{figure*}[t]
\centering
\includegraphics[width=0.75\linewidth]{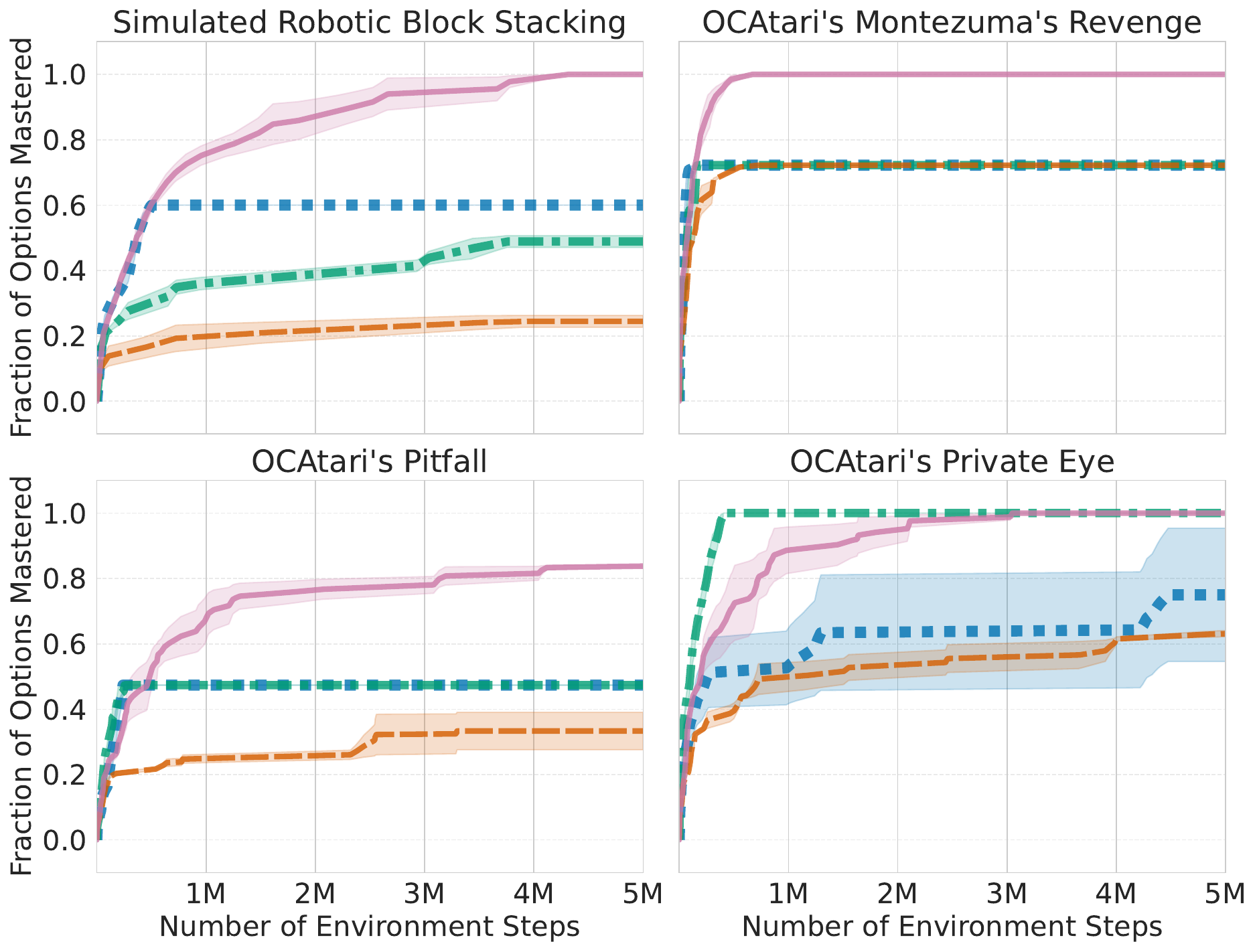}
\includegraphics[width=0.95\linewidth]{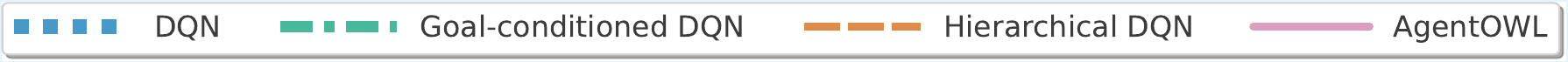}
\caption{Fraction of options mastered vs number of environment steps for the block stacking domain and the three OCAtari's games we test on: Montezuma's Revenge, Pitfall, and Private Eye. The shaded area represents standard error computed over 3 seeds. Option is acquired once its success rate for the recent episodes reaches threshold $\delta=0.5$.} 
\label{fig:main_results}
\end{figure*}

%% file: figs/ablation_results.tex
\begin{figure*}[t]
\centering
\includegraphics[width=\linewidth]{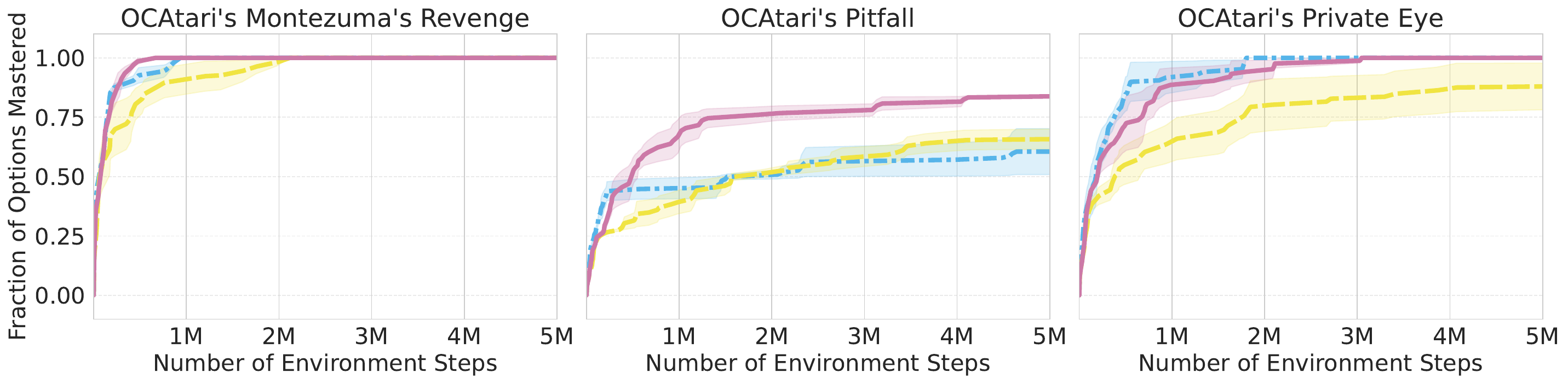}
\includegraphics[width=0.75\linewidth]{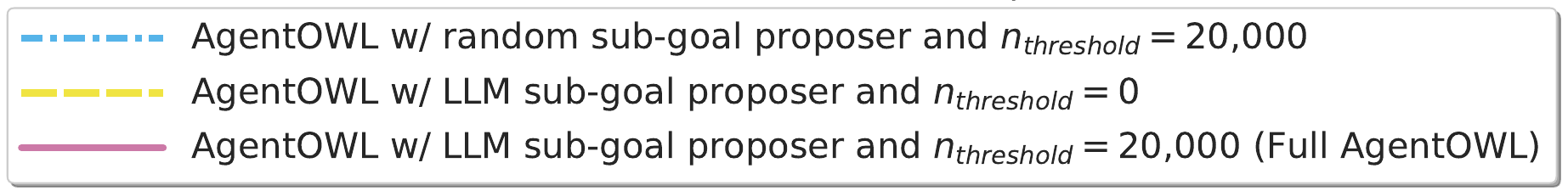}
\caption{Ablation study: Removing core components of AgentOWL lead to fewer options being acquired and/or more data from the environment being needed. Setting $n_{threshold} = 0$ means stabilization for hierarchical DQN is not implemented.}
\label{fig:ablate_results}
\end{figure*}

%% file: figs/implicit_learning.tex
\begin{figure*}[t]
\centering
\adjustbox{valign=t}{\includegraphics[width=0.35\linewidth]{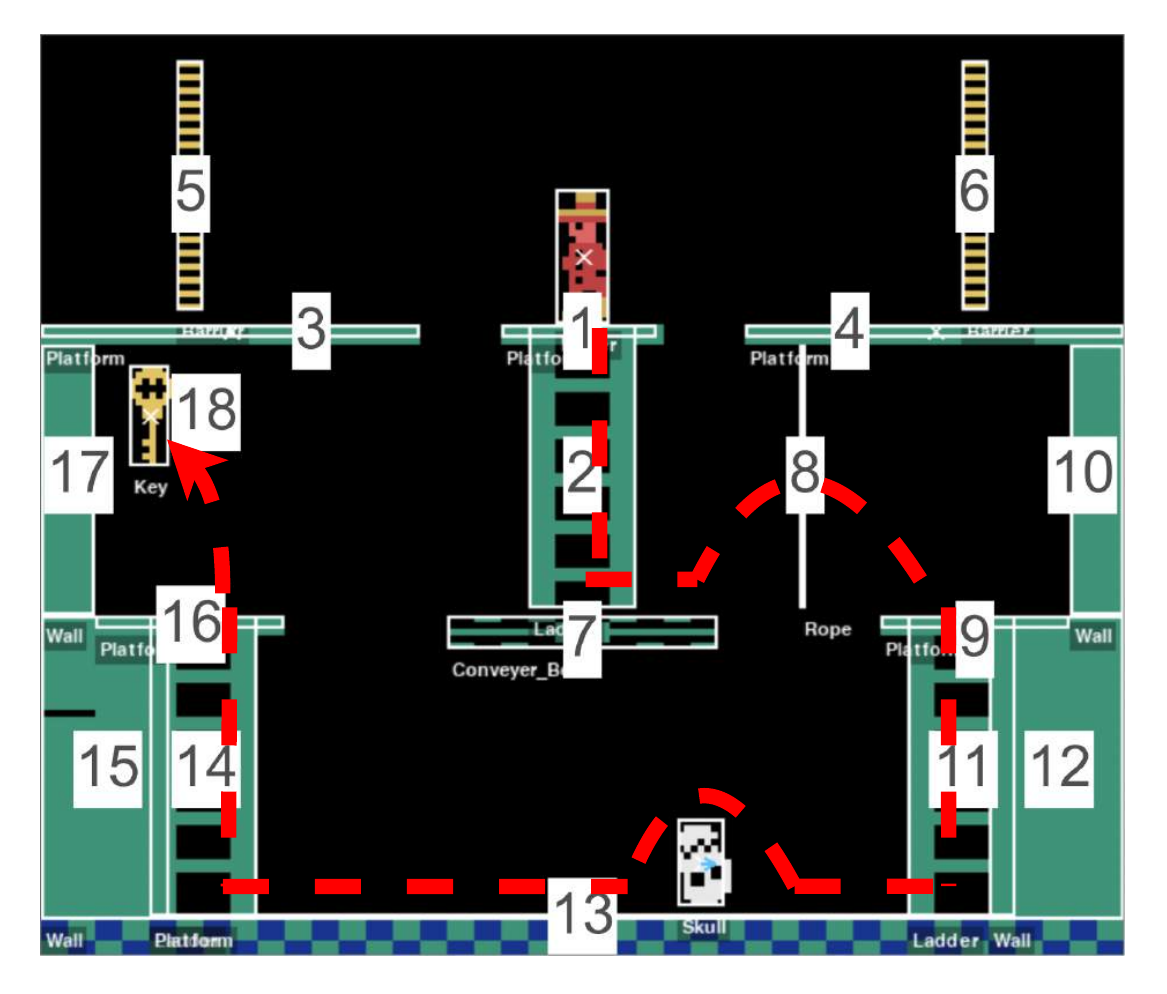}}
\adjustbox{valign=t}{\includegraphics[width=0.64\linewidth]{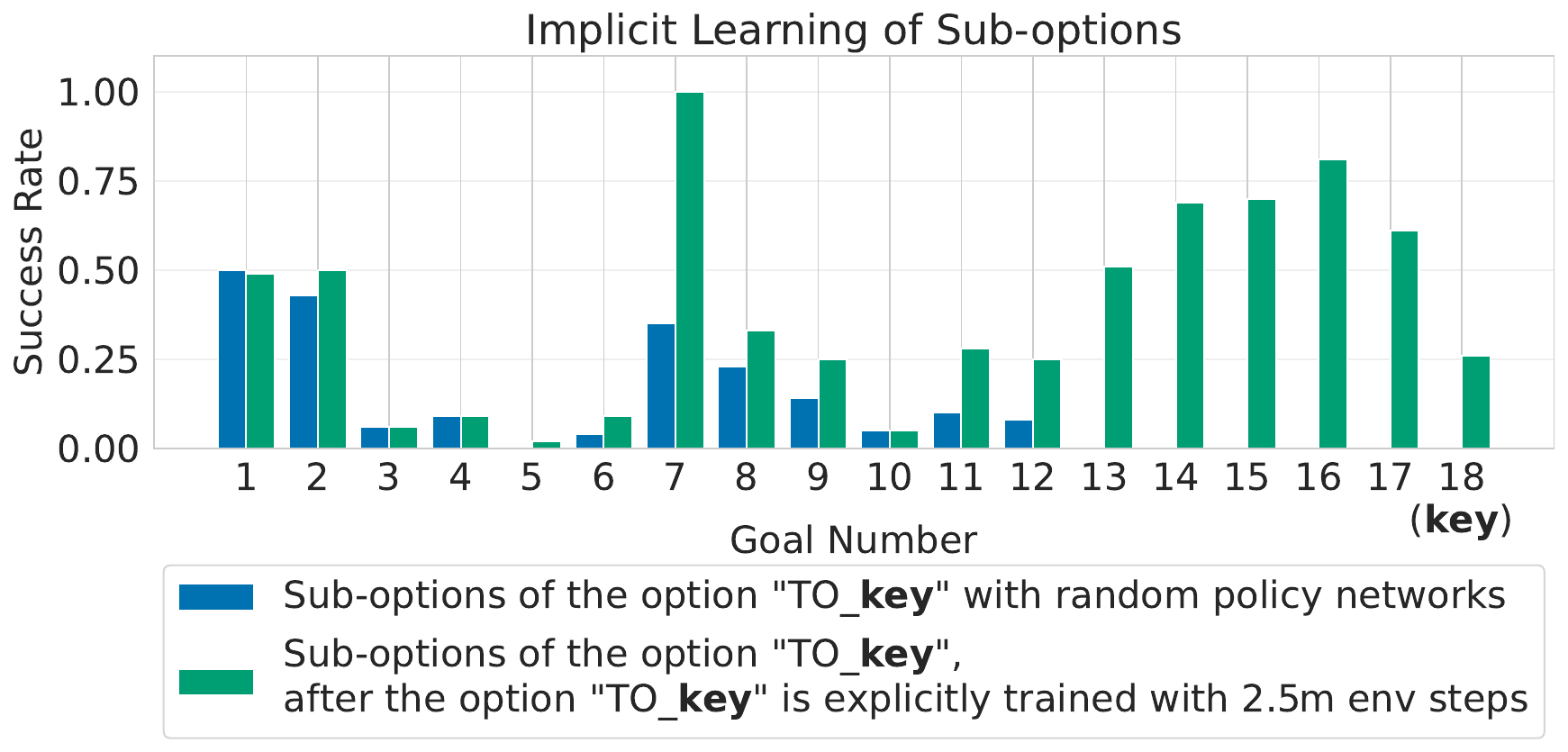}}
\caption{Left: Illustration of the goal labels (where all goals are of the form "touch this specific object") and the path the player needs to take to get the key from the starting position. 
Right: The success rates of the sub-options implicitly trained while the agent is training for the goal "key", compared to those of sub-options with randomly initialized policy network. 
The sub-options for the sub-goals within the path to retrieve the key, such as goal \#13-17, has high success rates, because they are useful to its parent-option "TO\_key".
On the other hand, unhelpful goals, such as goal \#3-6, remain poorly learned.
}
\label{fig:implicit_learning}
\end{figure*}

%% file: figs/zero_shot_generalization.tex
\begin{table}[b]
\vspace{1em}
\centering
\small
\begin{subtable}[b]{0.58\linewidth}
\centering
\begin{tabular}{lccc}
\toprule
Method & Goal $1$ & Goal $2$ & Goal $3$ \\
\midrule
Random & $0.06$ & $0.00$ & $0.00$ \\
Goal-conditioned DQN & $0.00$ & $0.00$ & $0.00$ \\
AgentOWL w/o new option & $0.02$ & $0.01$ & $0.26$ \\
AgentOWL w/ new option & $\mathbf{1.00}$ & $\mathbf{1.00}$ & $\mathbf{1.00}$ \\
\bottomrule
\end{tabular}
\caption{Private Eye}
\label{fig:zero_shot}
\end{subtable}\hfill
\begin{subtable}[b]{0.38\linewidth}
\centering
\setlength\tabcolsep{8pt}
\renewcommand{\arraystretch}{1.95}
\begin{tabular}{lcc}
\toprule
Method & \makecell{Seen\\layouts} & \makecell{Unseen\\layout} \\
\midrule
DQN & $0.96$ & $0.73$ \\
AgentOWL & $0.98$ & $\mathbf{1.00}$ \\
\bottomrule
\end{tabular}
\caption{Block stacking}
\label{tab:zero_shot_comp}
\end{subtable}
\caption{Zero-shot generalization results.
(a) Goal completion rates of previously mastered goals over 100 episodes in Private Eye, attempted from a new starting state. We then give a new option to navigate from the new starting state to the
initial starting state to AgentOWL, allowing it to achieve the goals from the new starting position without any additional training. \Cref{fig:gen} displays the initial state, the new starting state, and the goals in this experiment.
(b) Goal completion rate over 100 episodes for stacking the red block on the green block. Each method is trained on 15 starting layouts (seen) and tested on one held-out layout (unseen).
}
\end{table}


%% file: figs/dqn_pseudocode.tex
\begin{algorithm*}[!h]
\caption{DQN learning algorithm (without parallel environments)}
\small
\begin{algorithmic}[1]
\STATE Given an environment $E$, a list of actions/sub-options $\Omega$, , a reward function $R$, and training hyperparameters in \cref{tab:hyperparameters}.
\STATE Initialy policy network: $\pi \gets InitPolicy(E, \Omega)$
\STATE Initialize replay buffer $D \gets \{\}$
\STATE Get initial state: $\vs \gets E.reset()$
\FOR{$t \gets 0$ to $total\_env\_steps$ step $train\_frequency$}
\STATE \textit{\# Collect rollouts by executing $\pi$ for $train\_frequency$ steps}
\FOR{$i \gets 0$ to $train\_frequency$} 
\STATE Sample action: $a \sim \pi(a\mid\vs)$
\STATE Interact with environment: $\vs' \gets E.step(a)$
\STATE Update replay buffer: $D \gets D \cup \{(\vs,a,\vs', \kappa \cdot R(\vs, a, \vs')\}$
\STATE Update current state $\vs \gets \vs'$

\ENDFOR
\STATE \textit{\# Then, train for $n\_gradient\_steps$ steps}
\STATE Optimize the policy weights: $DQNUpdate(\pi, D, n\_gradient\_steps)$
\ENDFOR
\STATE \textbf{return} $\pi$
\end{algorithmic}
\label{fig:dqn_pseudocode}
\end{algorithm*}

%% file: figs/hdqn_pseudocode.tex
\begin{algorithm*}[!t]
\caption{Stable Hierarchical DQN learning algorithm (without parallel environments)\\
Note that ExecuteOneStep and ReceiveObsOneStep are defined in \Cref{fig:helper_pseudocode} and \Cref{fig:helper_pseudocode_2} respectively}
\small
\begin{algorithmic}[1]
\STATE Given $E, \Omega, D_{\Omega}, \pi_\Omega, \pi_g,g,n,\delta$ and training hyperparameters in \cref{tab:hyperparameters} and \cref{tab:hyperparameters_hdqn}.
\STATE
\STATE \textit{\# We define a \textbf{mutable} struct to keep track of execution variables of an option}
\STATE \textit{\# In hierarchical options, each parent option selects a child option to execute, recursing until reaching a primitive action}
\STATE \textbf{struct} $OptionExecutionState$
\STATE \quad Option: $o$
\STATE \quad Policy: $\pi$
\STATE \quad Goal: $g$
\STATE \quad Active child option: $child$
\STATE \quad Child start state: $u$
\STATE \quad Episode data: $D$
\STATE \quad Current execution time: $t$
\STATE \quad New data counter: $ct$
\STATE \textbf{end struct}
\STATE
\STATE Initialize (mutable) option execution states: \\ \;\;\;\; $\omega_\Omega \gets \{OptionExecutionState(o, \pi_o, g_o, None, None, [], 0, 0)\}_{o \in \Omega}$ 
\STATE Initialize (mutable) target option execution state: \\ \;\;\;\; $\omega_{target} \gets OptionExecutionState(o_g, \pi_g, g, None, None, [], 0, 0)$
\STATE Get initial state: $s \gets E.reset()$
\FOR{$i \gets 0$ to $total\_env\_steps$} 
\STATE Execute the target option for one step: $a \gets ExecuteOneStep(s, \omega_{target},\Omega, \omega_\Omega)$ 
\STATE Interact with environment: $s' \gets E.step(a)$
\STATE Update option states and replay buffers: $(\_, D_\Omega)\gets ReceiveObsOneStep(s', \omega_{target}, \Omega, D_\Omega, n, \delta)$ 
\FOR{$o \in \Omega$}
\STATE \textit{\# optimize the policy weights if the option has collected enough new data}
\IF{enough new data collected: $\omega_o.ct > train\_frequency$}
\STATE Optimize the policy weights: $DQNUpdate(\omega_o.\pi, D_o, n\_steps=n\_steps\_per\_sample \cdot \omega_o.ct)$
\STATE Reset new data counter: $\omega_o.ct \gets 0$ 
\ENDIF
\ENDFOR
\ENDFOR
\STATE Retrieve $\pi_\Omega$ from the mutable execution states: $\pi_\Omega \gets \{\omega_o.\pi \}_{o \in \Omega}$
\STATE Retrieve $\pi_g$ from mutable target execution state: $\pi_g \gets \omega_{target}.\pi$
\STATE \textbf{return} $\pi_g, \pi_\Omega, D_{\Omega}$
\end{algorithmic}
\label{fig:hdqn_pseudocode}
\end{algorithm*}

%% file: figs/hdqn_pseudocode_2.tex
\begin{algorithm*}[!t]
\caption{Helper function $ExecuteOneStep$ used in Stable Hierarchical DQN}
\small
\begin{algorithmic}[1]
\FUNCTION{$ExecuteOneStep(s, \omega, \Omega, \omega_\Omega)$}
\STATE \textit{\# keep recursing until we hit the leaf node of the execution trace}
\IF{$\omega.child \neq None$}
\STATE \textbf{return} $ExecuteOneStep(s, \omega.child,\Omega)$
\ENDIF
\STATE 
\STATE Sample a child option to execute: $a \gets \omega.\pi(a\mid s)$
\STATE Set child start state: $\omega.u \gets s$
\STATE Increment the number of sub-options $\omega$ has executed: $\omega.t \gets \omega.t + 1$
\STATE 

\STATE \textit{\# recurse if $a$ is indeed a sub-option, not a primitive action}
\IF{$a \in \Omega$}
\STATE Make sub-option $a$ (along with its execution state) the active child: $\omega.child \gets \omega_a$
\STATE \textbf{return} $ExecuteOneStep(s, \omega.child, \Omega)$
\ENDIF
\STATE 
\STATE \textit{\# return $a$ if it is a primitive action}
\STATE \textbf{return} $a$
\ENDFUNCTION
\end{algorithmic}
\label{fig:helper_pseudocode}
\end{algorithm*}

%% file: figs/hdqn_pseudocode_3.tex
\begin{algorithm*}[!t]
\caption{Helper function $ReceiveObsOneStep$ used in Stable Hierarchical DQN}
\small
\begin{algorithmic}[1]



\FUNCTION{$ReceiveObsOneStep(s', \omega, \Omega, D_\Omega, n, \delta)$}
\STATE \textit{\# keep recursing until we hit the leaf node of the execution trace}
\IF{$\omega.child \neq None$}
\STATE $(is\_child\_done, D_\Omega) \gets ReceiveObsOneStep(s', \omega.child)$
\ELSE
\STATE $is\_child\_done \gets True$
\ENDIF

\STATE 
\STATE \textit{\# Add data to episode buffer if the child option is done}
\IF{$is\_child\_done$}
\STATE Get child start state: $s \gets \omega.u$
\STATE Get child option: $a \gets \omega.child.o$
\STATE Calculate reward: $r \gets R_{\omega.g}(s,a,s')$
\STATE Add datapoint to $\omega.D \gets \omega.D \cup \{(s,a,s',r)\}$
\STATE Child is no longer active: $\omega.child \gets None$
\ENDIF
\STATE 
\STATE Check if the current option is done: $is\_cur\_done \gets (\omega.g(s') = 1) \text{ or } (\omega.t > max\_t)$
\STATE
\STATE \textit{\# Update execution state and potentially update the option's replay buffer if the option is done executing}
\IF{$is\_cur\_done$}
\STATE
\STATE \textit{\# Check if all executed sub-options in the episode are stable}
\STATE $only\_stable\_in\_episode \gets True$
\FOR{$(\_,a,\_,\_) \in \omega.D$}
\STATE \textit{\# $a$ is not stable if $a$ is a sub-option}
\STATE \textit{\# AND has lower number of samples it has been trained with than $n$ and lower goal completion rate than $\delta$}
\IF{$a \in \Omega$ and $n_a < n$ and $\delta_a < \delta$}
\STATE $only\_stable\_in\_episode \gets False$
\ENDIF
\ENDFOR
\STATE 
\STATE \textit{\# Add episode buffer to the full replay buffer if all executed sub-options in the episode are stable}
\IF{$only\_stable\_in\_episode$}
\STATE Get option index: $o \gets \omega.o$
\STATE Add episode data to the corresponding replay buffer: $D_o \gets D_o \cup \omega.D$
\STATE Add number of new data to the counter: $\omega.ct \gets \omega.ct + |\omega.D|$
\STATE Retrieve current epsilon: $(\pi^{real}, \pi^{wm}, \epsilon) \gets \omega.\pi$
\STATE Anneal $\epsilon$ based on the number of new data: $\epsilon \gets Anneal(\epsilon,|\omega.D|)$ 
\STATE Update the policy with the new $\epsilon$: $\omega.\pi \gets (\pi^{real}, \pi^{wm}, \epsilon)$
\ENDIF
\STATE 
\STATE \textit{\# Update execution state}
\STATE Child is no longer active: $\omega.child \gets None$
\STATE Clear episode buffer: $\omega.D \gets \{\}$
\STATE Execution time resets: $\omega.t \gets 0$
\ENDIF
\STATE 
\STATE \textbf{return} $is\_cur\_done, D_\Omega$
\ENDFUNCTION
\end{algorithmic}
\label{fig:helper_pseudocode_2}
\end{algorithm*}

%% file: figs/hyperparameters.tex
\begin{table}[t]
    \centering
\centering
\setlength\tabcolsep{4pt} 

\small{
\begin{tabular}{c@{\hspace{4pt}}c@{\hspace{3pt}}}
\toprule
Hyperparameter & Value \\ 
\midrule
Number of parallel environments & $4$\\
Batch size & $256$\\
Learning rate & $0.0001$\\
Replay buffer size & $5000 \cdot 4$\\
Multi-step return & $10$\\
Discount factor $\gamma$ & $0.99$\\
Priority replay temperature & $0.01$\\
Target network update interval & $200$\\
Number of gradient steps per training step & $64$\\
Training frequency & $16 \cdot 4$\\
Reward multiplier $\kappa$ & $10$\\
Maximum gradient norm & $10$\\
Random exploration rate $\epsilon$ & $0$\\
Frame stacking & $4$\\
Optimizer & Kron \cite{castanyer2025stable,li2017preconditioned}\\
\bottomrule\\
\end{tabular}}
\caption{Common training hyperparameters for Rainbow DQN.}
\label{tab:hyperparameters}
\end{table}

%% file: figs/hyperparameters_hdqn.tex
\begin{table}[t]
    \centering
\centering
\setlength\tabcolsep{4pt} 

\small{
\begin{tabular}{c@{\hspace{4pt}}c@{\hspace{3pt}}}
\toprule
Hyperparameter & Value \\ 
\midrule
Max option executime time $max\_t$ & $100$ \\
Gradient steps per sample $n\_steps\_per\_sample$ & $1$\\
Annealing schedule $\epsilon$ & $Linear(start = 1, end = 0, n\_samples=10000)$\\
$n_{threshold}$ & $20000$\\
$\delta_{threshold}$ & $0.5$\\
\bottomrule\\
\end{tabular}}
\caption{Training hyperparameters that are specific to Hierarchical DQN.}
\label{tab:hyperparameters_hdqn}
\end{table}

%% file: figs/hyperparameters_wm.tex
\begin{table}[t]
    \centering
\centering
\setlength\tabcolsep{4pt} 

\small{
\begin{tabular}{c@{\hspace{4pt}}c@{\hspace{3pt}}}
\toprule
Hyperparameter & Value \\ 
\midrule
Multi-step return & $1$\\
Random exploration rate $\epsilon$  & $Linear(start=1.0, end=0.01, fraction=0.95)$ \\

\bottomrule\\
\end{tabular}}
\caption{Training hyperparameters for Rainbow DQN that are specific to $\pi^{wm}$.}
\label{tab:hyperparameters_wm}
\end{table}

%% file: figs/gen.tex
\begin{figure*}[t]
\centering
\includegraphics[width=\linewidth]{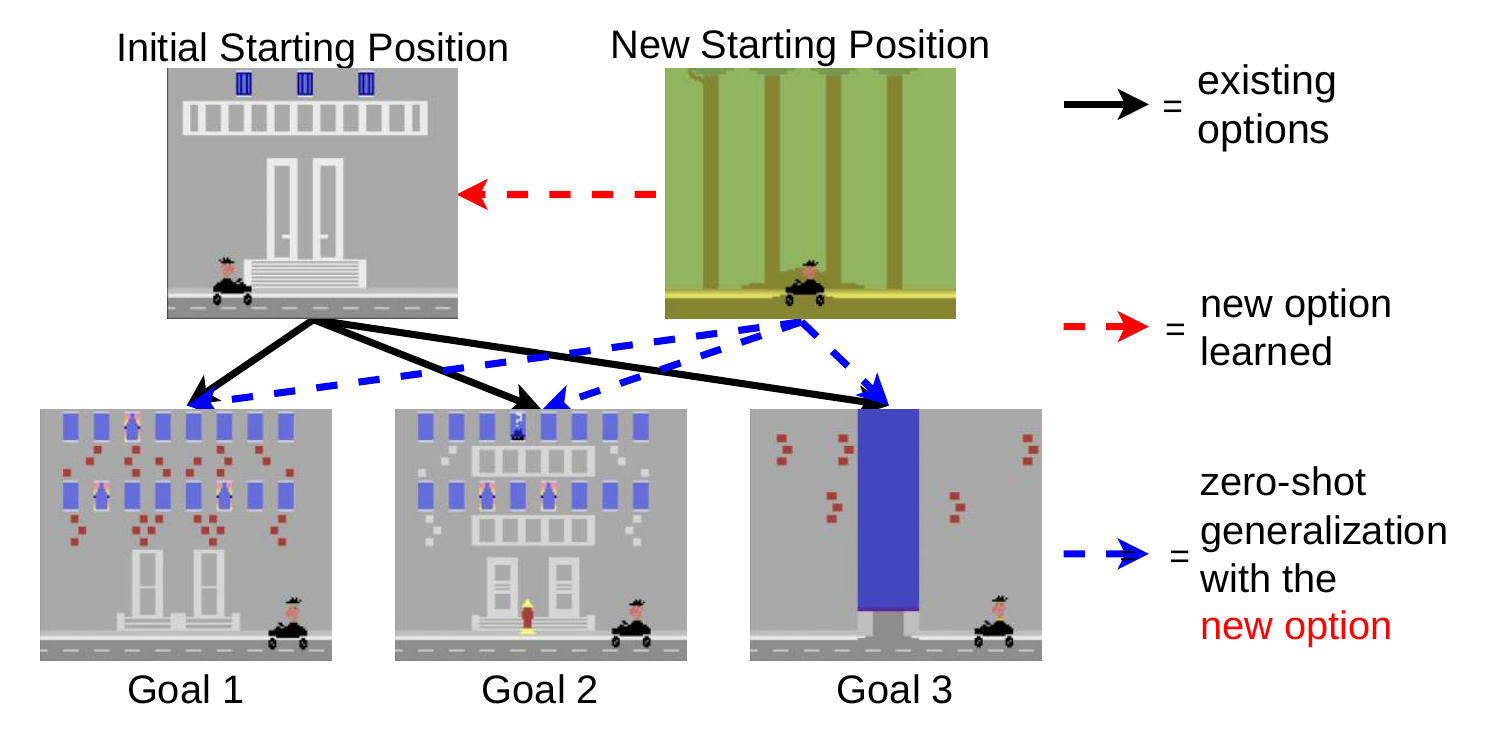}
\caption{Illustration of the initial state, the new starting state, and the goals in \cref{fig:zero_shot}.} 
\label{fig:gen}
\end{figure*}

%% file: figs/mr_goals.tex
\begin{figure*}[t]
\centering
\includegraphics[width=0.4\linewidth]{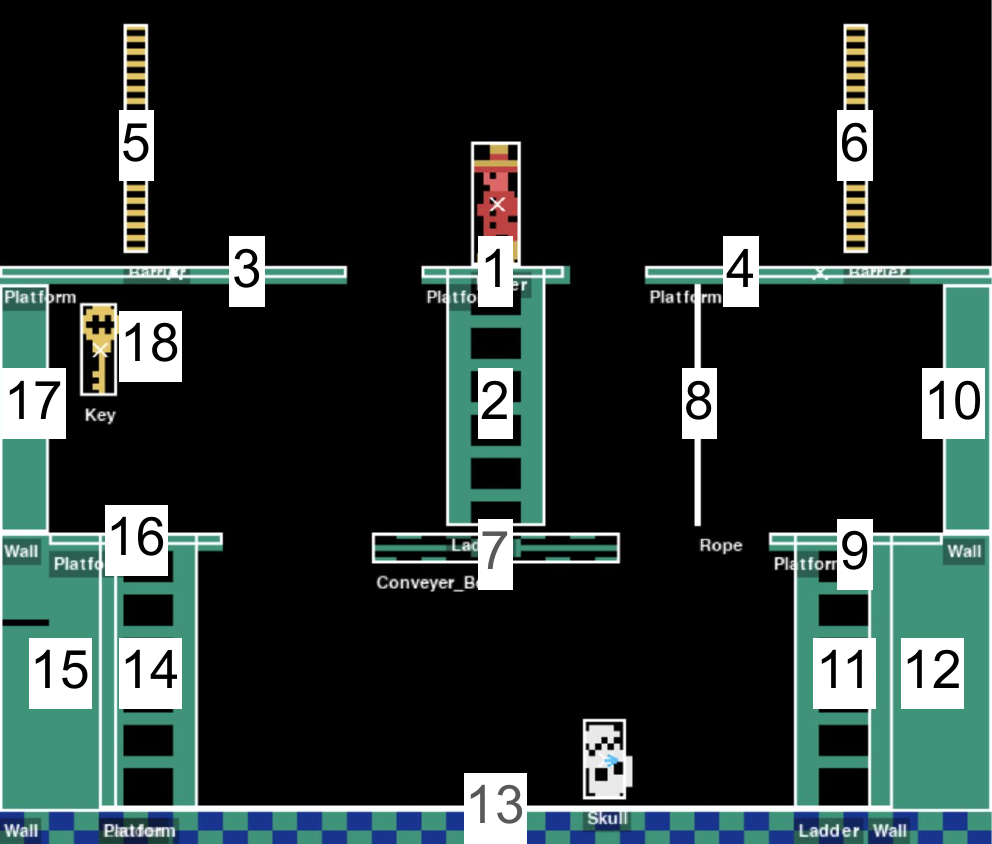}
\caption{Screenshots of Montezuma's Revenge with goals labeled in order it appears in the ordered list of goals} 
\label{fig:mr_goals}
\end{figure*}

%% file: figs/pitfall_goals.tex
\begin{figure*}[t]
\centering
\includegraphics[width=\linewidth]{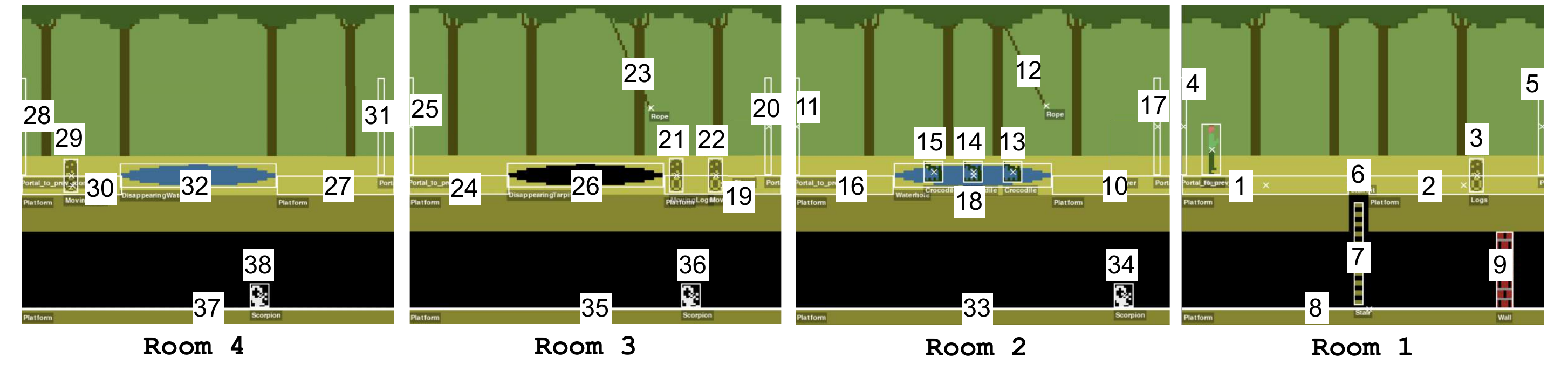}
\caption{Screenshots of Pitfall with goals labeled in order it appears in the ordered list of goals} 
\label{fig:pitfall_goals}
\end{figure*}

%% file: figs/privateeye_goals.tex
\begin{figure*}[t]
\centering
\includegraphics[width=\linewidth]{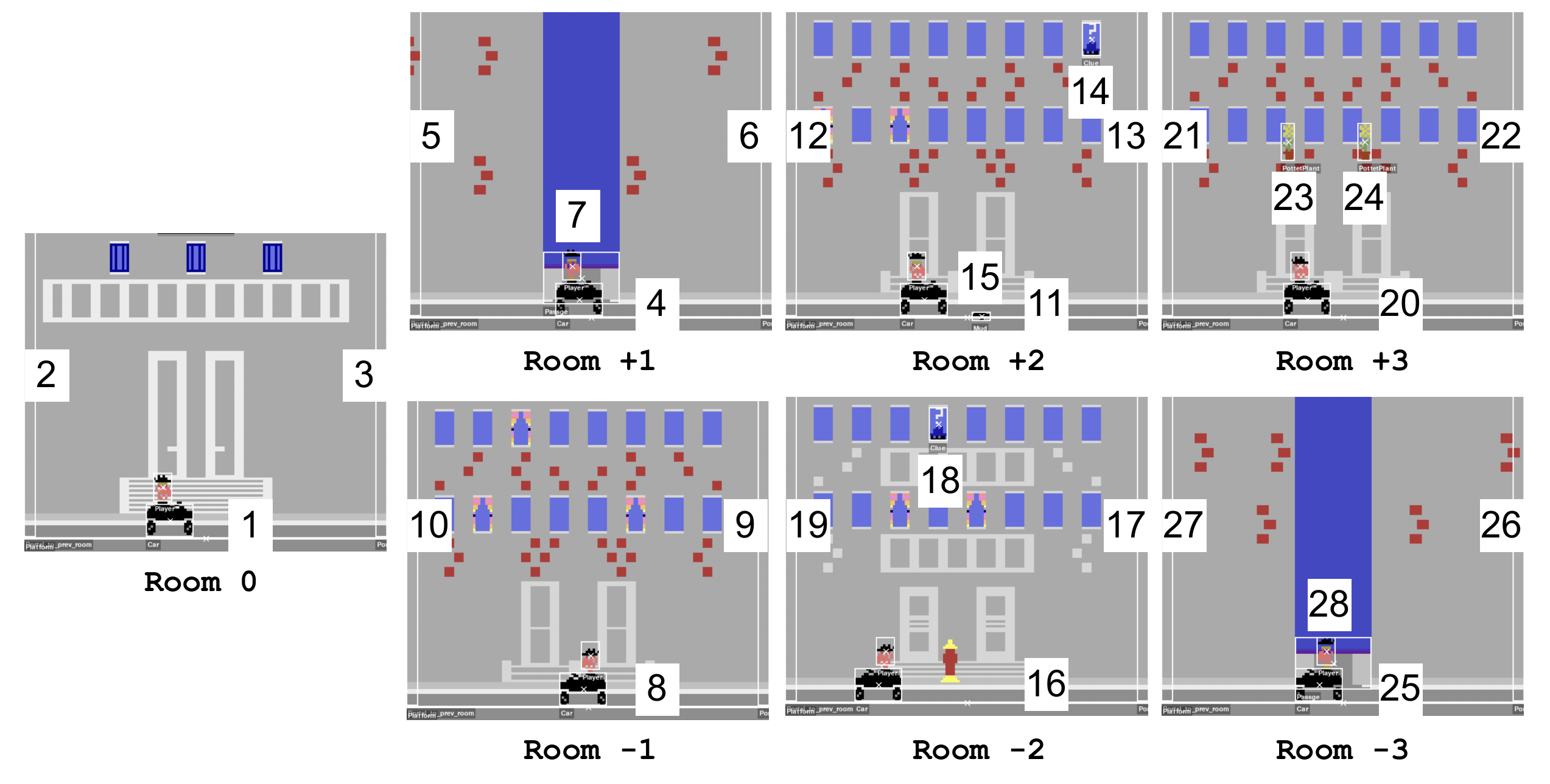}
\caption{Screenshots of Private Eye with goals labeled in order it appears in the ordered list of goals} 
\label{fig:privateeye_goals}
\end{figure*}

%% file: prompts/subgoal_proposer.tex
\begin{table}[t]
\centering
\setlength\tabcolsep{4pt} 
\begin{tabular}{@{}c@{}}
\toprule
\begin{minipage}{\linewidth}

\small{
\begin{spverbatim}
Here is the current observation of the game {game_name}:

{cur_obs}

And here is a list of goals we know how to achieve:

{achieved_goal_names_and_descriptions}

Your task is to list 1 achieved goals that, on its own, can act as a possible stepping stone to achieve the target goal of '{target_goal_name}' -- Description: '{target_goal_description}'.

Required reasoning process:
First, discuss out loud how to achieve the target goal of '{target_goal_name}', 
taken into account the current observation.
Then, for each achieved goal, discuss out loud how completing that goal would help us achieve the target goal of '{target_goal_name}', taken into account the current observation.
Make sure to go through all achieved goals. But also do not keep repeating the same achieved goal. The current position of the player is irrelevant.

Final output format:
After you are done reasoning, list the achieved goal in a numbered list with 
the following format:
Possible stepping stone 1: <achieved goal>
\end{spverbatim}}
\end{minipage}
\\
\bottomrule
\\
\end{tabular}
\caption{Prompt for LLM to propose sub-goals to hypothesize new sub-options. Note that $target\_goal\_description$ is simply ``touch object with id $obj\_id$ or ``in room \#$room\_num$, touch object with id $obj\_id$'' and $target\_goal\_name$ is something like ``bot\_plat'' or ``rope\_in\_roomnumber+2''. $cur\_obs$ is a state sampled from $D_{seen}$ if there is a single room like Montezuma's Revenge and multiple states, one per room, for Pitfall and Private Eye.}
\label{prompt:subgoal}
\end{table}

%% file: prompts/poe_world_1.tex
\begin{table}[t]
\centering
\setlength\tabcolsep{4pt} 
\begin{tabular}{@{}c@{}}
\toprule
\begin{minipage}{\linewidth}

\small{
\begin{spverbatim}
I'll give you an input list of objects. 

I want you to list 4 possible features that the input list of objects has that allows us to achieve a certain goal.

Here's an example:

Example input list of objects:
player object with at (x=16, y=104, w=8, h=21),
wall object with at (x=136, y=148, w=7, h=32),
logs object with at (x=125, y=118, w=6, h=14),
stairpit object with at (x=76, y=122, w=8, h=6),
stair object with at (x=78, y=136, w=4, h=42),
platform object with at (x=8, y=179, w=152, h=1),
platform object with at (x=8, y=125, w=152, h=1),
playerscore object with at (x=38, y=9, w=30, h=8),
lifecount object with at (x=23, y=22, w=1, h=8),
lifecount object with at (x=21, y=22, w=1, h=8),
timer object with at (x=31, y=22, w=37, h=8),
portal_0 object with at (x=7, y=85, w=1, h=40),
portal_1 object with at (x=155, y=85, w=1, h=40),
Interaction -- player object with at (x=16, y=104, w=8, h=21) is touching platform object with at (x=8, y=125, w=152, h=1)

Example possible features that allow us to achieve the goal of '{goal}':
1. AnyObjTypeTouching: The player object touches a platform object
2. SpecificObjTouching: The player object touches the platform object located at (x=8, y=125)
3. AnyObjTypeTouching: ...
4. SpecificObjTouching: ...

Now, I want you to list 4 possible features of the input list of objects has that allows us to achieve the goal of '{goal}'.

Input list of objects:
{input}

Please follow these rules for your output:
1. Do not explain -- simply list each feature
2. Make the features diverse
3. Do use interactions (what the player is touching), as they usually make good features
4. Each rule should of type 'AnyObjTypeTouching' or 'SpecificObjTouching'
\end{spverbatim}}
\end{minipage}
\\
\bottomrule
\\
\end{tabular}
\caption{Prompt for LLM to propose preconditions for games where the agent controls only the Player object: Montezuma's Revenge and Pitfall.}
\label{prompt:poe_world_1}
\end{table}

%% file: prompts/poe_world_2.tex
\begin{table}[t]
\centering
\setlength\tabcolsep{4pt} 
\begin{tabular}{@{}c@{}}
\toprule
\begin{minipage}{\linewidth}

\small{
\begin{spverbatim}
I'll give you an input list of objects. 

I want you to list 4 possible features that the input list of objects has that allows us to achieve a certain goal.

Here's an example:

Example input list of objects:
player object (x=30, y=150, w=8, h=12),
car object (x=27, y=163, w=20, h=14),
score object (x=75, y=8, w=30, h=8),
clock object (x=67, y=19, w=30, h=8),
roomnumber_+0 object (x=0, y=0, w=0, h=0),
portal_to_prev_room object (x=8, y=27, w=5, h=150),
portal_to_next_room object (x=155, y=27, w=5, h=150),
platform object (x=8, y=177, w=152, h=1),

Example possible features that allow us to achieve the goal of '{goal}':
1. RoomNumberExist: An object with type 'roomnumber_+0' exists
2. ObjTouchingAndRoomNumberExist: The car object touches the platform object and an object with type 'roomnumber_+0' exists

Now, I want you to list 2 possible features of the input list of objects has that allows us to achieve the goal of '{goal}'.

Input list of objects:
{input}

Please follow these rules for your output:
1. Do not explain -- simply list each feature
2. Each rule should of type 'RoomNumberExist' or 'ObjTouchingAndRoomNumberExist'
3. Make sure to mention the roomnumber in the feature, e.g., 'an object with type 'roomnumber_+0' exists'
\end{spverbatim}}
\end{minipage}
\\
\bottomrule
\\
\end{tabular}
\caption{Prompt for LLM to propose preconditions for games where the agent controls several objects: Private Eye (the agent controls Player and Car object)}
\label{prompt:poe_world_2}
\end{table}

%% file: checklist.tex
\section*{NeurIPS Paper Checklist}

\begin{enumerate}

\item {\bf Claims}
    \item[] Question: Do the main claims made in the abstract and introduction accurately reflect the paper's contributions and scope?
    \item[] Answer: \answerYes{} 
    \item[] Justification: The claims made in the abstract and introduction accurately reflect the paper’s
contributions and scope.
    \item[] Guidelines:
    \begin{itemize}
        \item The answer \answerNA{} means that the abstract and introduction do not include the claims made in the paper.
        \item The abstract and/or introduction should clearly state the claims made, including the contributions made in the paper and important assumptions and limitations. A \answerNo{} or \answerNA{} answer to this question will not be perceived well by the reviewers. 
        \item The claims made should match theoretical and experimental results, and reflect how much the results can be expected to generalize to other settings. 
        \item It is fine to include aspirational goals as motivation as long as it is clear that these goals are not attained by the paper. 
    \end{itemize}

\item {\bf Limitations}
    \item[] Question: Does the paper discuss the limitations of the work performed by the authors?
    \item[] Answer: \answerYes{} 
    \item[] Justification: The paper discusses the limitations of the work performed by the authors.
    \item[] Guidelines:
    \begin{itemize}
        \item The answer \answerNA{} means that the paper has no limitation while the answer \answerNo{} means that the paper has limitations, but those are not discussed in the paper. 
        \item The authors are encouraged to create a separate ``Limitations'' section in their paper.
        \item The paper should point out any strong assumptions and how robust the results are to violations of these assumptions (e.g., independence assumptions, noiseless settings, model well-specification, asymptotic approximations only holding locally). The authors should reflect on how these assumptions might be violated in practice and what the implications would be.
        \item The authors should reflect on the scope of the claims made, e.g., if the approach was only tested on a few datasets or with a few runs. In general, empirical results often depend on implicit assumptions, which should be articulated.
        \item The authors should reflect on the factors that influence the performance of the approach. For example, a facial recognition algorithm may perform poorly when image resolution is low or images are taken in low lighting. Or a speech-to-text system might not be used reliably to provide closed captions for online lectures because it fails to handle technical jargon.
        \item The authors should discuss the computational efficiency of the proposed algorithms and how they scale with dataset size.
        \item If applicable, the authors should discuss possible limitations of their approach to address problems of privacy and fairness.
        \item While the authors might fear that complete honesty about limitations might be used by reviewers as grounds for rejection, a worse outcome might be that reviewers discover limitations that aren't acknowledged in the paper. The authors should use their best judgment and recognize that individual actions in favor of transparency play an important role in developing norms that preserve the integrity of the community. Reviewers will be specifically instructed to not penalize honesty concerning limitations.
    \end{itemize}

\item {\bf Theory assumptions and proofs}
    \item[] Question: For each theoretical result, does the paper provide the full set of assumptions and a complete (and correct) proof?
    \item[] Answer: \answerNA{} 
    \item[] Justification: The paper does not include theoretical results.
    \item[] Guidelines:
    \begin{itemize}
        \item The answer \answerNA{} means that the paper does not include theoretical results. 
        \item All the theorems, formulas, and proofs in the paper should be numbered and cross-referenced.
        \item All assumptions should be clearly stated or referenced in the statement of any theorems.
        \item The proofs can either appear in the main paper or the supplemental material, but if they appear in the supplemental material, the authors are encouraged to provide a short proof sketch to provide intuition. 
        \item Inversely, any informal proof provided in the core of the paper should be complemented by formal proofs provided in appendix or supplemental material.
        \item Theorems and Lemmas that the proof relies upon should be properly referenced. 
    \end{itemize}

    \item {\bf Experimental result reproducibility}
    \item[] Question: Does the paper fully disclose all the information needed to reproduce the main experimental results of the paper to the extent that it affects the main claims and/or conclusions of the paper (regardless of whether the code and data are provided or not)?
    \item[] Answer: \answerYes{} 
    \item[] Justification: We have fully described the algorithms used in the paper, with details included
in the appendix.
    \item[] Guidelines:
    \begin{itemize}
        \item The answer \answerNA{} means that the paper does not include experiments.
        \item If the paper includes experiments, a \answerNo{} answer to this question will not be perceived well by the reviewers: Making the paper reproducible is important, regardless of whether the code and data are provided or not.
        \item If the contribution is a dataset and\slash or model, the authors should describe the steps taken to make their results reproducible or verifiable. 
        \item Depending on the contribution, reproducibility can be accomplished in various ways. For example, if the contribution is a novel architecture, describing the architecture fully might suffice, or if the contribution is a specific model and empirical evaluation, it may be necessary to either make it possible for others to replicate the model with the same dataset, or provide access to the model. In general. releasing code and data is often one good way to accomplish this, but reproducibility can also be provided via detailed instructions for how to replicate the results, access to a hosted model (e.g., in the case of a large language model), releasing of a model checkpoint, or other means that are appropriate to the research performed.
        \item While NeurIPS does not require releasing code, the conference does require all submissions to provide some reasonable avenue for reproducibility, which may depend on the nature of the contribution. For example
        \begin{enumerate}
            \item If the contribution is primarily a new algorithm, the paper should make it clear how to reproduce that algorithm.
            \item If the contribution is primarily a new model architecture, the paper should describe the architecture clearly and fully.
            \item If the contribution is a new model (e.g., a large language model), then there should either be a way to access this model for reproducing the results or a way to reproduce the model (e.g., with an open-source dataset or instructions for how to construct the dataset).
            \item We recognize that reproducibility may be tricky in some cases, in which case authors are welcome to describe the particular way they provide for reproducibility. In the case of closed-source models, it may be that access to the model is limited in some way (e.g., to registered users), but it should be possible for other researchers to have some path to reproducing or verifying the results.
        \end{enumerate}
    \end{itemize}

\item {\bf Open access to data and code}
    \item[] Question: Does the paper provide open access to the data and code, with sufficient instructions to faithfully reproduce the main experimental results, as described in supplemental material?
    \item[] Answer: \answerNo{} 
    \item[] Justification: We currently cannot release code and data, but we have fully described the algorithms used in the paper, with details included
in the appendix.
    \item[] Guidelines:
    \begin{itemize}
        \item The answer \answerNA{} means that paper does not include experiments requiring code.
        \item Please see the NeurIPS code and data submission guidelines (\url{https://neurips.cc/public/guides/CodeSubmissionPolicy}) for more details.
        \item While we encourage the release of code and data, we understand that this might not be possible, so \answerNo{} is an acceptable answer. Papers cannot be rejected simply for not including code, unless this is central to the contribution (e.g., for a new open-source benchmark).
        \item The instructions should contain the exact command and environment needed to run to reproduce the results. See the NeurIPS code and data submission guidelines (\url{https://neurips.cc/public/guides/CodeSubmissionPolicy}) for more details.
        \item The authors should provide instructions on data access and preparation, including how to access the raw data, preprocessed data, intermediate data, and generated data, etc.
        \item The authors should provide scripts to reproduce all experimental results for the new proposed method and baselines. If only a subset of experiments are reproducible, they should state which ones are omitted from the script and why.
        \item At submission time, to preserve anonymity, the authors should release anonymized versions (if applicable).
        \item Providing as much information as possible in supplemental material (appended to the paper) is recommended, but including URLs to data and code is permitted.
    \end{itemize}

\item {\bf Experimental setting/details}
    \item[] Question: Does the paper specify all the training and test details (e.g., data splits, hyperparameters, how they were chosen, type of optimizer) necessary to understand the results?
    \item[] Answer: \answerYes{} 
    \item[] Justification: Full experimental details are provided in appendix.
    \item[] Guidelines:
    \begin{itemize}
        \item The answer \answerNA{} means that the paper does not include experiments.
        \item The experimental setting should be presented in the core of the paper to a level of detail that is necessary to appreciate the results and make sense of them.
        \item The full details can be provided either with the code, in appendix, or as supplemental material.
    \end{itemize}

\item {\bf Experiment statistical significance}
    \item[] Question: Does the paper report error bars suitably and correctly defined or other appropriate information about the statistical significance of the experiments?
    \item[] Answer: \answerYes{} 
    \item[] Justification: Our main results are accompanied by error bars.
    \item[] Guidelines:
    \begin{itemize}
        \item The answer \answerNA{} means that the paper does not include experiments.
        \item The authors should answer \answerYes{} if the results are accompanied by error bars, confidence intervals, or statistical significance tests, at least for the experiments that support the main claims of the paper.
        \item The factors of variability that the error bars are capturing should be clearly stated (for example, train/test split, initialization, random drawing of some parameter, or overall run with given experimental conditions).
        \item The method for calculating the error bars should be explained (closed form formula, call to a library function, bootstrap, etc.)
        \item The assumptions made should be given (e.g., Normally distributed errors).
        \item It should be clear whether the error bar is the standard deviation or the standard error of the mean.
        \item It is OK to report 1-sigma error bars, but one should state it. The authors should preferably report a 2-sigma error bar than state that they have a 96\% CI, if the hypothesis of Normality of errors is not verified.
        \item For asymmetric distributions, the authors should be careful not to show in tables or figures symmetric error bars that would yield results that are out of range (e.g., negative error rates).
        \item If error bars are reported in tables or plots, the authors should explain in the text how they were calculated and reference the corresponding figures or tables in the text.
    \end{itemize}

\item {\bf Experiments compute resources}
    \item[] Question: For each experiment, does the paper provide sufficient information on the computer resources (type of compute workers, memory, time of execution) needed to reproduce the experiments?
    \item[] Answer: \answerYes{} 
    \item[] Justification: We have provided compute resources
used for our experiments.
    \item[] Guidelines:
    \begin{itemize}
        \item The answer \answerNA{} means that the paper does not include experiments.
        \item The paper should indicate the type of compute workers CPU or GPU, internal cluster, or cloud provider, including relevant memory and storage.
        \item The paper should provide the amount of compute required for each of the individual experimental runs as well as estimate the total compute. 
        \item The paper should disclose whether the full research project required more compute than the experiments reported in the paper (e.g., preliminary or failed experiments that didn't make it into the paper). 
    \end{itemize}
    
\item {\bf Code of ethics}
    \item[] Question: Does the research conducted in the paper conform, in every respect, with the NeurIPS Code of Ethics \url{https://neurips.cc/public/EthicsGuidelines}?
    \item[] Answer: \answerYes{} 
    \item[] Justification: The research conducted in the paper conforms, in every respect, with the NeurIPS Code of Ethics. 
    \item[] Guidelines:
    \begin{itemize}
        \item The answer \answerNA{} means that the authors have not reviewed the NeurIPS Code of Ethics.
        \item If the authors answer \answerNo, they should explain the special circumstances that require a deviation from the Code of Ethics.
        \item The authors should make sure to preserve anonymity (e.g., if there is a special consideration due to laws or regulations in their jurisdiction).
    \end{itemize}

\item {\bf Broader impacts}
    \item[] Question: Does the paper discuss both potential positive societal impacts and negative societal impacts of the work performed?
    \item[] Answer: \answerNA{} 
    \item[] Justification: Our work is foundational research and has no direct societal impact.
    \item[] Guidelines:
    \begin{itemize}
        \item The answer \answerNA{} means that there is no societal impact of the work performed.
        \item If the authors answer \answerNA{} or \answerNo, they should explain why their work has no societal impact or why the paper does not address societal impact.
        \item Examples of negative societal impacts include potential malicious or unintended uses (e.g., disinformation, generating fake profiles, surveillance), fairness considerations (e.g., deployment of technologies that could make decisions that unfairly impact specific groups), privacy considerations, and security considerations.
        \item The conference expects that many papers will be foundational research and not tied to particular applications, let alone deployments. However, if there is a direct path to any negative applications, the authors should point it out. For example, it is legitimate to point out that an improvement in the quality of generative models could be used to generate Deepfakes for disinformation. On the other hand, it is not needed to point out that a generic algorithm for optimizing neural networks could enable people to train models that generate Deepfakes faster.
        \item The authors should consider possible harms that could arise when the technology is being used as intended and functioning correctly, harms that could arise when the technology is being used as intended but gives incorrect results, and harms following from (intentional or unintentional) misuse of the technology.
        \item If there are negative societal impacts, the authors could also discuss possible mitigation strategies (e.g., gated release of models, providing defenses in addition to attacks, mechanisms for monitoring misuse, mechanisms to monitor how a system learns from feedback over time, improving the efficiency and accessibility of ML).
    \end{itemize}
    
\item {\bf Safeguards}
    \item[] Question: Does the paper describe safeguards that have been put in place for responsible release of data or models that have a high risk for misuse (e.g., pre-trained language models, image generators, or scraped datasets)?
    \item[] Answer: \answerNA{} 
    \item[] Justification: The paper poses no such risks.
    \item[] Guidelines:
    \begin{itemize}
        \item The answer \answerNA{} means that the paper poses no such risks.
        \item Released models that have a high risk for misuse or dual-use should be released with necessary safeguards to allow for controlled use of the model, for example by requiring that users adhere to usage guidelines or restrictions to access the model or implementing safety filters. 
        \item Datasets that have been scraped from the Internet could pose safety risks. The authors should describe how they avoided releasing unsafe images.
        \item We recognize that providing effective safeguards is challenging, and many papers do not require this, but we encourage authors to take this into account and make a best faith effort.
    \end{itemize}

\item {\bf Licenses for existing assets}
    \item[] Question: Are the creators or original owners of assets (e.g., code, data, models), used in the paper, properly credited and are the license and terms of use explicitly mentioned and properly respected?
    \item[] Answer: \answerYes{} 
    \item[] Justification: The creators of original data are all properly credited, and the license and
terms of use of the data are explicitly mentioned and properly respected.
    \item[] Guidelines:
    \begin{itemize}
        \item The answer \answerNA{} means that the paper does not use existing assets.
        \item The authors should cite the original paper that produced the code package or dataset.
        \item The authors should state which version of the asset is used and, if possible, include a URL.
        \item The name of the license (e.g., CC-BY 4.0) should be included for each asset.
        \item For scraped data from a particular source (e.g., website), the copyright and terms of service of that source should be provided.
        \item If assets are released, the license, copyright information, and terms of use in the package should be provided. For popular datasets, \url{paperswithcode.com/datasets} has curated licenses for some datasets. Their licensing guide can help determine the license of a dataset.
        \item For existing datasets that are re-packaged, both the original license and the license of the derived asset (if it has changed) should be provided.
        \item If this information is not available online, the authors are encouraged to reach out to the asset's creators.
    \end{itemize}

\item {\bf New assets}
    \item[] Question: Are new assets introduced in the paper well documented and is the documentation provided alongside the assets?
    \item[] Answer: \answerNA{} 
    \item[] Justification: The paper does not release new assets.
    \item[] Guidelines:
    \begin{itemize}
        \item The answer \answerNA{} means that the paper does not release new assets.
        \item Researchers should communicate the details of the dataset\slash code\slash model as part of their submissions via structured templates. This includes details about training, license, limitations, etc. 
        \item The paper should discuss whether and how consent was obtained from people whose asset is used.
        \item At submission time, remember to anonymize your assets (if applicable). You can either create an anonymized URL or include an anonymized zip file.
    \end{itemize}

\item {\bf Crowdsourcing and research with human subjects}
    \item[] Question: For crowdsourcing experiments and research with human subjects, does the paper include the full text of instructions given to participants and screenshots, if applicable, as well as details about compensation (if any)? 
    \item[] Answer: \answerNA{} 
    \item[] Justification: The paper does not involve crowdsourcing nor research with human subjects.
    \item[] Guidelines:
    \begin{itemize}
        \item The answer \answerNA{} means that the paper does not involve crowdsourcing nor research with human subjects.
        \item Including this information in the supplemental material is fine, but if the main contribution of the paper involves human subjects, then as much detail as possible should be included in the main paper. 
        \item According to the NeurIPS Code of Ethics, workers involved in data collection, curation, or other labor should be paid at least the minimum wage in the country of the data collector. 
    \end{itemize}

\item {\bf Institutional review board (IRB) approvals or equivalent for research with human subjects}
    \item[] Question: Does the paper describe potential risks incurred by study participants, whether such risks were disclosed to the subjects, and whether Institutional Review Board (IRB) approvals (or an equivalent approval/review based on the requirements of your country or institution) were obtained?
    \item[] Answer: \answerNA{} 
    \item[] Justification: The paper does not involve crowdsourcing nor research with human subjects.
    \item[] Guidelines:
    \begin{itemize}
        \item The answer \answerNA{} means that the paper does not involve crowdsourcing nor research with human subjects.
        \item Depending on the country in which research is conducted, IRB approval (or equivalent) may be required for any human subjects research. If you obtained IRB approval, you should clearly state this in the paper. 
        \item We recognize that the procedures for this may vary significantly between institutions and locations, and we expect authors to adhere to the NeurIPS Code of Ethics and the guidelines for their institution. 
        \item For initial submissions, do not include any information that would break anonymity (if applicable), such as the institution conducting the review.
    \end{itemize}

\item {\bf Declaration of LLM usage}
    \item[] Question: Does the paper describe the usage of LLMs if it is an important, original, or non-standard component of the core methods in this research? Note that if the LLM is used only for writing, editing, or formatting purposes and does \emph{not} impact the core methodology, scientific rigor, or originality of the research, declaration is not required.
    \item[] Answer: \answerYes{} 
    \item[] Justification: This paper describes the usage of LLMs in our method.
    \item[] Guidelines:
    \begin{itemize}
        \item The answer \answerNA{} means that the core method development in this research does not involve LLMs as any important, original, or non-standard components.
        \item Please refer to our LLM policy in the NeurIPS handbook for what should or should not be described.
    \end{itemize}

\end{enumerate}

%% file: example_paper.bib
@inproceedings{tang2024worldcoder,
title={WorldCoder, a Model-Based {LLM} Agent: Building World Models by Writing Code and Interacting with the Environment},
author={Hao Tang and Darren Yan Key and Kevin Ellis},
booktitle={The Thirty-eighth Annual Conference on Neural Information Processing Systems},
year={2024},
url={https://openreview.net/forum?id=QGJSXMhVaL}
}

@article{sutton1999between,
  title={Between MDPs and semi-MDPs: A framework for temporal abstraction in reinforcement learning},
  author={Sutton, Richard S and Precup, Doina and Singh, Satinder},
  journal={Artificial intelligence},
  volume={112},
  number={1-2},
  pages={181--211},
  year={1999},
  publisher={Elsevier}
}

@inproceedings{bengio2009curriculum,
  title={Curriculum learning},
  author={Bengio, Yoshua and Louradour, J{\'e}r{\^o}me and Collobert, Ronan and Weston, Jason},
  booktitle={ICML},
  year={2009}
}

@article{mnih2015human,
  title={Human-level control through deep reinforcement learning},
  author={Volodymyr Mnih and Koray Kavukcuoglu and David Silver and Andrei A. Rusu and Joel Veness and Marc G. Bellemare and Alex Graves and Martin A. Riedmiller and Andreas Kirkeby Fidjeland and Georg Ostrovski and Stig Petersen and Charlie Beattie and Amir Sadik and Ioannis Antonoglou and Helen King and Dharshan Kumaran and Daan Wierstra and Shane Legg and Demis Hassabis},
  journal={Nature},
  volume={518},
  number={7540},
  pages={529--533},
  year={2015},
  publisher={Nature Publishing Group}
}

@article{wang2023voyager,
  title   = {Voyager: An Open-Ended Embodied Agent with Large Language Models},
  author  = {Guanzhi Wang and Yuqi Xie and Yunfan Jiang and Ajay Mandlekar and Chaowei Xiao and Yuke Zhu and Linxi Fan and Anima Anandkumar},
  year    = {2023},
  journal = {arXiv preprint arXiv: Arxiv-2305.16291}
}

@article{yao2022react,
  title={ReAct: Synergizing Reasoning and Acting in Language Models},
  author={Yao, Shunyu and Zhao, Jeffrey and Yu, Dian and Du, Nan and Shafran, Izhak and Narasimhan, Karthik and Cao, Yuan},
  journal={arXiv preprint arXiv:2210.03629},
  year={2022}
}

@article{delfosse2023ocatari,
  title={Ocatari: Object-centric atari 2600 reinforcement learning environments},
  author={Delfosse, Quentin and Bl{\"u}ml, Jannis and Gregori, Bjarne and Sztwiertnia, Sebastian and Kersting, Kristian},
  journal={arXiv preprint arXiv:2306.08649},
  year={2023}
}

@Article{bellemare13atari,
    author = {{Bellemare}, M.~G. and {Naddaf}, Y. and {Veness}, J. and {Bowling}, M.},
    title = {The Arcade Learning Environment: An Evaluation Platform for General Agents},
    journal = {Journal of Artificial Intelligence Research},
    year = "2013",
    month = "jun",
    volume = "47",
    pages = "253--279",
}

@Article{machado18arcade,
    author = {Marlos C. Machado and Marc G. Bellemare and Erik Talvitie and Joel Veness and Matthew J. Hausknecht and Michael Bowling},
    title = {Revisiting the Arcade Learning Environment: Evaluation Protocols and Open Problems for General Agents},
    journal = {Journal of Artificial Intelligence Research},
    volume = {61},
    pages = {523--562},
    year = {2018}
}

@inproceedings{alonso2024diamond,
      title={Diffusion for World Modeling: Visual Details Matter in Atari},
      author={Eloi Alonso and Adam Jelley and Vincent Micheli and Anssi Kanervisto and Amos Storkey and Tim Pearce and François Fleuret},
      booktitle={Thirty-eighth Conference on Neural Information Processing Systems},
      year={2024},
      url={https://arxiv.org/abs/2405.12399},
}

@inproceedings{dainese2024codeworldmodel,
title={Generating Code World Models with Large Language Models Guided by Monte Carlo Tree Search}, 
author={Nicola Dainese and Matteo Merler and Minttu Alakuijala and Pekka Marttinen},
booktitle={The Thirty-eighth Annual Conference on Neural Information Processing Systems},
year={2024},
url={https://arxiv.org/abs/2405.15383}, 
}

@article{liang2024visualpredicator,
  title={Visualpredicator: Learning abstract world models with neuro-symbolic predicates for robot planning},
  author={Liang, Yichao and Kumar, Nishanth and Tang, Hao and Weller, Adrian and Tenenbaum, Joshua B and Silver, Tom and Henriques, Jo{\~a}o F and Ellis, Kevin},
  journal={arXiv preprint arXiv:2410.23156},
  year={2024}
}

@inproceedings{kaelbling2011hierarchical,
  title={Hierarchical task and motion planning in the now},
  author={Kaelbling, Leslie Pack and Lozano-P{\'e}rez, Tom{\'a}s},
  booktitle={2011 IEEE International Conference on Robotics and Automation},
  pages={1470--1477},
  year={2011},
  organization={IEEE}
}

@article{ahmed2025synthesizing,
  title={Synthesizing world models for bilevel planning},
  author={Ahmed, Zergham and Tenenbaum, Joshua B and Bates, Christopher J and Gershman, Samuel J},
  journal={arXiv preprint arXiv:2503.20124},
  year={2025}
}

@article{ecoffet2021first,
  title={First return, then explore},
  author={Ecoffet, Adrien and Huizinga, Joost and Lehman, Joel and Stanley, Kenneth O and Clune, Jeff},
  journal={Nature},
  volume={590},
  number={7847},
  pages={580--586},
  year={2021},
  publisher={Nature Publishing Group UK London}
}

@article{curtis2024partially,
  title={Partially observable task and motion planning with uncertainty and risk awareness},
  author={Curtis, Aidan and Matheos, George and Gothoskar, Nishad and Mansinghka, Vikash and Tenenbaum, Joshua and Lozano-P{\'e}rez, Tom{\'a}s and Kaelbling, Leslie Pack},
  journal={arXiv preprint arXiv:2403.10454},
  year={2024}
}

@inproceedings{bacon2017option,
  title={The option-critic architecture},
  author={Bacon, Pierre-Luc and Harb, Jean and Precup, Doina},
  booktitle={Proceedings of the AAAI conference on artificial intelligence},
  volume={31},
  number={1},
  year={2017}
}

@article{piriyakulkij2025poeworld,
    title={PoE-World: Compositional World Modeling with Products of Programmatic Experts}, 
    author={Wasu Top Piriyakulkij and Yichao Liang and Hao Tang and Adrian Weller and Marta Kryven and Kevin Ellis},
    year={2025},
    journal={Advances in Neural Information Processing Systems},
}

@article{khan2025one,
  title={One Life to Learn: Inferring Symbolic World Models for Stochastic Environments from Unguided Exploration},
  author={Khan, Zaid and Prasad, Archiki and Stengel-Eskin, Elias and Cho, Jaemin and Bansal, Mohit},
  journal={arXiv preprint arXiv:2510.12088},
  year={2025}
}

@Book{puterman1994mdp,
  author    = {Puterman, Martin L.},
  title     = {Markov Decision Processes: Discrete Stochastic Dynamic Programming},
  publisher = {John Wiley \& Sons},
  year      = {1994},
  address   = {New York},
  isbn      = {978-0471727828}
}

@article{khetarpal2021temporally,
  title={Temporally abstract partial models},
  author={Khetarpal, Khimya and Ahmed, Zafarali and Comanici, Gheorghe and Precup, Doina},
  journal={Advances in Neural Information Processing Systems},
  volume={34},
  pages={1979--1991},
  year={2021}
}

@article{nachum2018data,
  title={Data-efficient hierarchical reinforcement learning},
  author={Nachum, Ofir and Gu, Shixiang Shane and Lee, Honglak and Levine, Sergey},
  journal={Advances in neural information processing systems},
  volume={31},
  year={2018}
}

@inproceedings{hessel2018rainbow,
  title={Rainbow: Combining improvements in deep reinforcement learning},
  author={Hessel, Matteo and Modayil, Joseph and Van Hasselt, Hado and Schaul, Tom and Ostrovski, Georg and Dabney, Will and Horgan, Dan and Piot, Bilal and Azar, Mohammad and Silver, David},
  booktitle={Proceedings of the AAAI conference on artificial intelligence},
  volume={32},
  number={1},
  year={2018}
}

@article{mcdermott20001998,
  title={The 1998 AI planning systems competition},
  author={McDermott, Drew M},
  journal={AI magazine},
  volume={21},
  number={2},
  pages={35--35},
  year={2000}
}

@Book{bertsekas1995,
author = {Bertsekas, Dimitri},
year = {1995},
month = {01},
pages = {},
title = {Dynamic Programming and Optimal Control},
volume = {1},
publisher = {Athena Scientific},
}

@article{li2006towards,
  title={Towards a unified theory of state abstraction for MDPs.},
  author={Li, Lihong and Walsh, Thomas J and Littman, Michael L},
  journal={AI\&M},
  volume={1},
  number={2},
  pages={3},
  year={2006}
}

@article{ball2025genie,
  title={Genie 3: A new frontier for world models},
  author={Ball, Philip J and Bauer, Jakob and Belletti, Frank and Brownfield, B and Ephrat, A and Fruchter, S and Gupta, A and Holsheimer, K and Holynski, A and Hron, J and others},
  journal={Google DeepMind Blog},
  pages={253--279},
  year={2025}
}

@article{moerland2023model,
  title={Model-based reinforcement learning: A survey},
  author={M. Moerland, Thomas and Broekens, Joost and Plaat, Aske and M. Jonker, Catholijn},
  journal={Foundations and Trends in Machine Learning},
  volume={16},
  number={1},
  pages={1--118},
  year={2023},
  publisher={Emerald Publishing Limited}
}

@article{aytar2018playing,
  title={Playing hard exploration games by watching youtube},
  author={Aytar, Yusuf and Pfaff, Tobias and Budden, David and Paine, Thomas and Wang, Ziyu and De Freitas, Nando},
  journal={Advances in neural information processing systems},
  volume={31},
  year={2018}
}

@article{hosu2016playing,
  title={Playing atari games with deep reinforcement learning and human checkpoint replay},
  author={Hosu, Ionel-Alexandru and Rebedea, Traian},
  journal={arXiv preprint arXiv:1607.05077},
  year={2016}
}

@article{nica2022paradox,
  title={The paradox of choice: Using attention in hierarchical reinforcement learning},
  author={Nica, Andrei and Khetarpal, Khimya and Precup, Doina},
  journal={arXiv preprint arXiv:2201.09653},
  year={2022}
}

@inproceedings{johnson2016malmo,
  title={The Malmo Platform for Artificial Intelligence Experimentation.},
  author={Johnson, Matthew and Hofmann, Katja and Hutton, Tim and Bignell, David},
  booktitle={Ijcai},
  volume={16},
  pages={4246--4247},
  year={2016}
}

@article{kuttler2020nethack,
  title={The nethack learning environment},
  author={K{\"u}ttler, Heinrich and Nardelli, Nantas and Miller, Alexander and Raileanu, Roberta and Selvatici, Marco and Grefenstette, Edward and Rockt{\"a}schel, Tim},
  journal={Advances in Neural Information Processing Systems},
  volume={33},
  pages={7671--7684},
  year={2020}
}

@misc{sutton2025oak,
  author       = {Sutton, Richard S.},
  title        = {The Oak Architecture: A Vision of SuperIntelligence from Experience},
  howpublished = {Invited talk at the Conference on Neural Information Processing Systems (NeurIPS)},
  year         = {2025},
  month        = {December},
  address      = {Mexico City, Mexico},
  url          = {https://neurips.cc/virtual/2025/loc/mexico-city/invited-talk/129132}
}

@article{konidaris2009skill,
  title={Skill discovery in continuous reinforcement learning domains using skill chaining},
  author={Konidaris, George and Barto, Andrew},
  journal={Advances in neural information processing systems},
  volume={22},
  year={2009}
}

@article{klissarovdoro2023motif,
    title={Motif: Intrinsic Motivation From Artificial Intelligence Feedback},
    author={Klissarov, Martin and D’Oro, Pierluca and Sodhani, Shagun and Raileanu, Roberta and Bacon, Pierre-Luc and Vincent, Pascal and Zhang, Amy and Henaff, Mikael},
    year={2023},
    month={9},
    journal={arXiv preprint arXiv:2310.00166}
}

@inproceedings{wong2024ada,
  title = {Learning Adaptive Planning Representations with Natural Language Guidance},
  author = {Lio Wong and Jiayuan Mao and Pratyusha Sharma and Zachary S. Siegel and Jiahai Feng and Noa Korneev and Joshua B. Tenenbaum and Jacob Andreas},
  booktitle = {International Conference on Learning Representations (ICLR)},
  year = {2024}
}

@article{lehrach2025code,
  title={Code world models for general game playing},
  author={Lehrach, Wolfgang and Hennes, Daniel and Lazaro-Gredilla, Miguel and Lou, Xinghua and Wendelken, Carter and Li, Zun and Dedieu, Antoine and Grau-Moya, Jordi and Lanctot, Marc and Iscen, Atil and others},
  journal={ICLR},
  year={2026}
}

@inproceedings{vezhnevets2017feudal,
  title={Feudal networks for hierarchical reinforcement learning},
  author={Vezhnevets, Alexander Sasha and Osindero, Simon and Schaul, Tom and Heess, Nicolas and Jaderberg, Max and Silver, David and Kavukcuoglu, Koray},
  booktitle={International conference on machine learning},
  pages={3540--3549},
  year={2017},
  organization={PMLR}
}

@article{kulkarni2016hierarchical,
  title={Hierarchical deep reinforcement learning: Integrating temporal abstraction and intrinsic motivation},
  author={Kulkarni, Tejas D and Narasimhan, Karthik and Saeedi, Ardavan and Tenenbaum, Josh},
  journal={Advances in neural information processing systems},
  volume={29},
  year={2016}
}

@inproceedings{nottingham2023embodied,
  title={Do embodied agents dream of pixelated sheep: Embodied decision making using language guided world modelling},
  author={Nottingham, Kolby and Ammanabrolu, Prithviraj and Suhr, Alane and Choi, Yejin and Hajishirzi, Hannaneh and Singh, Sameer and Fox, Roy},
  booktitle={International Conference on Machine Learning},
  pages={26311--26325},
  year={2023},
  organization={PMLR}
}

@inproceedings{khetarpal2020can,
  title={What can I do here? A theory of affordances in reinforcement learning},
  author={Khetarpal, Khimya and Ahmed, Zafarali and Comanici, Gheorghe and Abel, David and Precup, Doina},
  booktitle={International Conference on Machine Learning},
  pages={5243--5253},
  year={2020},
  organization={PMLR}
}

@article{kamat2020diversity,
  title={Diversity-enriched option-critic},
  author={Kamat, Anand and Precup, Doina},
  journal={arXiv preprint arXiv:2011.02565},
  year={2020}
}

@inproceedings{abdulhai2022context,
  title={Context-specific representation abstraction for deep option learning},
  author={Abdulhai, Marwa and Kim, Dong-Ki and Riemer, Matthew and Liu, Miao and Tesauro, Gerald and How, Jonathan P},
  booktitle={Proceedings of the AAAI Conference on Artificial Intelligence},
  volume={36},
  number={6},
  pages={5959--5967},
  year={2022}
}

@inproceedings{dean1997model,
  title={Model minimization in Markov decision processes},
  author={Dean, Thomas and Givan, Robert},
  booktitle={AAAI/IAAI},
  pages={106--111},
  year={1997}
}

@article{kaelbling1996reinforcement,
  title={Reinforcement learning: A survey},
  author={Kaelbling, Leslie Pack and Littman, Michael L and Moore, Andrew W},
  journal={Journal of artificial intelligence research},
  volume={4},
  pages={237--285},
  year={1996}
}

@article{eysenbach2018diversity,
  title={Diversity is all you need: Learning skills without a reward function},
  author={Eysenbach, Benjamin and Gupta, Abhishek and Ibarz, Julian and Levine, Sergey},
  journal={arXiv preprint arXiv:1802.06070},
  year={2018}
}

@book{parr1998hierarchical,
  title={Hierarchical control and learning for Markov decision processes},
  author={Parr, Ronald Edward},
  year={1998},
  publisher={University of California, Berkeley}
}

@article{dietterich2000hierarchical,
  title={Hierarchical reinforcement learning with the MAXQ value function decomposition},
  author={Dietterich, Thomas G},
  journal={Journal of artificial intelligence research},
  volume={13},
  pages={227--303},
  year={2000}
}

@article{liang2025exopredicator,
  title={ExoPredicator: Learning Abstract Models of Dynamic Worlds for Robot Planning},
  author={Liang, Yichao and Nguyen, Dat and Yang, Cambridge and Li, Tianyang and Tenenbaum, Joshua B and Rasmussen, Carl Edward and Weller, Adrian and Tavares, Zenna and Silver, Tom and Ellis, Kevin},
  journal={arXiv preprint arXiv:2509.26255},
  year={2025}
}

@article{athalye2024pixels,
  title={From Pixels to Predicates: Learning Symbolic World Models via Pretrained Vision-Language Models},
  author={Athalye, Ashay and Kumar, Nishanth and Silver, Tom and Liang, Yichao and Wang, Jiuguang and Lozano-P{\'e}rez, Tom{\'a}s and Kaelbling, Leslie Pack},
  journal={arXiv preprint arXiv:2501.00296},
  year={2024}
}

@article{fikes1971strips,
  title={STRIPS: A new approach to the application of theorem proving to problem solving},
  author={Fikes, Richard E and Nilsson, Nils J},
  journal={Artificial intelligence},
  volume={2},
  number={3-4},
  pages={189--208},
  year={1971},
  publisher={Elsevier}
}

@inproceedings{bagaria2019option,
  title={Option discovery using deep skill chaining},
  author={Bagaria, Akhil and Konidaris, George},
  booktitle={International Conference on Learning Representations},
  year={2019}
}

@inproceedings{bagaria2021robustly,
  title={Robustly learning composable options in deep reinforcement learning},
  author={Bagaria, Akhil and Senthil, Jason and Slivinski, Matthew and Konidaris, George},
  booktitle={Proceedings of the 30th International Joint Conference on Artificial Intelligence},
  year={2021}
}

@article{harutyunyan2019termination,
  title={The termination critic},
  author={Harutyunyan, Anna and Dabney, Will and Borsa, Diana and Heess, Nicolas and Munos, Remi and Precup, Doina},
  journal={arXiv preprint arXiv:1902.09996},
  year={2019}
}

@inproceedings{tiwari2019natural,
  title={Natural option critic},
  author={Tiwari, Saket and Thomas, Philip S},
  booktitle={Proceedings of the AAAI Conference on Artificial Intelligence},
  volume={33},
  number={01},
  pages={5175--5182},
  year={2019}
}

@article{dayan1992feudal,
  title={Feudal reinforcement learning},
  author={Dayan, Peter and Hinton, Geoffrey E},
  journal={Advances in neural information processing systems},
  volume={5},
  year={1992}
}

@inproceedings{bagaria2021skill,
  title={Skill discovery for exploration and planning using deep skill graphs},
  author={Bagaria, Akhil and Senthil, Jason K and Konidaris, George},
  booktitle={International conference on machine learning},
  pages={521--531},
  year={2021},
  organization={PMLR}
}

@article{florensa2017stochastic,
  title={Stochastic neural networks for hierarchical reinforcement learning},
  author={Florensa, Carlos and Duan, Yan and Abbeel, Pieter},
  journal={arXiv preprint arXiv:1704.03012},
  year={2017}
}

@article{heess2016learning,
  title={Learning and transfer of modulated locomotor controllers},
  author={Heess, Nicolas and Wayne, Greg and Tassa, Yuval and Lillicrap, Timothy and Riedmiller, Martin and Silver, David},
  journal={arXiv preprint arXiv:1610.05182},
  year={2016}
}

@article{li2019hierarchical,
  title={Hierarchical reinforcement learning with advantage-based auxiliary rewards},
  author={Li, Siyuan and Wang, Rui and Tang, Minxue and Zhang, Chongjie},
  journal={Advances in Neural Information Processing Systems},
  volume={32},
  year={2019}
}

@article{hafner2022deep,
  title={Deep hierarchical planning from pixels},
  author={Hafner, Danijar and Lee, Kuang-Huei and Fischer, Ian and Abbeel, Pieter},
  journal={Advances in Neural Information Processing Systems},
  volume={35},
  pages={26091--26104},
  year={2022}
}

@article{ma2023eureka,
  title={Eureka: Human-level reward design via coding large language models},
  author={Ma, Yecheng Jason and Liang, William and Wang, Guanzhi and Huang, De-An and Bastani, Osbert and Jayaraman, Dinesh and Zhu, Yuke and Fan, Linxi and Anandkumar, Anima},
  journal={arXiv preprint arXiv:2310.12931},
  year={2023}
}

@article{kwon2023reward,
  title={Reward design with language models},
  author={Kwon, Minae and Xie, Sang Michael and Bullard, Kalesha and Sadigh, Dorsa},
  journal={arXiv preprint arXiv:2303.00001},
  year={2023}
}

@article{ahn2022can,
  title={Do as i can, not as i say: Grounding language in robotic affordances},
  author={Ahn, Michael and Brohan, Anthony and Brown, Noah and Chebotar, Yevgen and Cortes, Omar and David, Byron and Finn, Chelsea and Fu, Chuyuan and Gopalakrishnan, Keerthana and Hausman, Karol and others},
  journal={arXiv preprint arXiv:2204.01691},
  year={2022}
}

@inproceedings{huang2022language,
  title={Language models as zero-shot planners: Extracting actionable knowledge for embodied agents},
  author={Huang, Wenlong and Abbeel, Pieter and Pathak, Deepak and Mordatch, Igor},
  booktitle={International conference on machine learning},
  pages={9118--9147},
  year={2022},
  organization={PMLR}
}

@article{singh2022progprompt,
  title={Progprompt: Generating situated robot task plans using large language models},
  author={Singh, Ishika and Blukis, Valts and Mousavian, Arsalan and Goyal, Ankit and Xu, Danfei and Tremblay, Jonathan and Fox, Dieter and Thomason, Jesse and Garg, Animesh},
  journal={arXiv preprint arXiv:2209.11302},
  year={2022}
}

@inproceedings{song2023llm,
  title={Llm-planner: Few-shot grounded planning for embodied agents with large language models},
  author={Song, Chan Hee and Wu, Jiaman and Washington, Clayton and Sadler, Brian M and Chao, Wei-Lun and Su, Yu},
  booktitle={Proceedings of the IEEE/CVF international conference on computer vision},
  pages={2998--3009},
  year={2023}
}

@article{liang2022code,
  title={Code as policies: Language model programs for embodied control},
  author={Liang, Jacky and Huang, Wenlong and Xia, Fei and Xu, Peng and Hausman, Karol and Ichter, Brian and Florence, Pete and Zeng, Andy},
  journal={arXiv preprint arXiv:2209.07753},
  year={2022}
}

@inproceedings{gu2016continuous,
  title={Continuous deep q-learning with model-based acceleration},
  author={Gu, Shixiang and Lillicrap, Timothy and Sutskever, Ilya and Levine, Sergey},
  booktitle={International conference on machine learning},
  pages={2829--2838},
  year={2016},
  organization={PMLR}
}

@article{janner2019trust,
  title={When to trust your model: Model-based policy optimization},
  author={Janner, Michael and Fu, Justin and Zhang, Marvin and Levine, Sergey},
  journal={Advances in neural information processing systems},
  volume={32},
  year={2019}
}

@article{asadi2019combating,
  title={Combating the compounding-error problem with a multi-step model},
  author={Asadi, Kavosh and Misra, Dipendra and Kim, Seungchan and Littman, Michel L},
  journal={arXiv preprint arXiv:1905.13320},
  year={2019}
}

@article{ba2016layer,
  title={Layer normalization},
  author={Ba, Jimmy Lei and Kiros, Jamie Ryan and Hinton, Geoffrey E},
  journal={arXiv preprint arXiv:1607.06450},
  year={2016}
}

@inproceedings{lyle2023understanding,
  title={Understanding plasticity in neural networks},
  author={Lyle, Clare and Zheng, Zeyu and Nikishin, Evgenii and Pires, Bernardo Avila and Pascanu, Razvan and Dabney, Will},
  booktitle={International Conference on Machine Learning},
  pages={23190--23211},
  year={2023},
  organization={PMLR}
}

@article{lyle2024disentangling,
  title={Disentangling the causes of plasticity loss in neural networks},
  author={Lyle, Clare and Zheng, Zeyu and Khetarpal, Khimya and van Hasselt, Hado and Pascanu, Razvan and Martens, James and Dabney, Will},
  journal={arXiv preprint arXiv:2402.18762},
  year={2024}
}

@inproceedings{wang2016dueling,
  title={Dueling network architectures for deep reinforcement learning},
  author={Wang, Ziyu and Schaul, Tom and Hessel, Matteo and Hasselt, Hado and Lanctot, Marc and Freitas, Nando},
  booktitle={International conference on machine learning},
  pages={1995--2003},
  year={2016},
  organization={PMLR}
}

@inproceedings{
fortunato2018noisy,
title={Noisy Networks For Exploration},
author={Meire Fortunato and Mohammad Gheshlaghi Azar and Bilal Piot and Jacob Menick and Matteo Hessel and Ian Osband and Alex Graves and Volodymyr Mnih and Remi Munos and Demis Hassabis and Olivier Pietquin and Charles Blundell and Shane Legg},
booktitle={International Conference on Learning Representations},
year={2018},
url={https://openreview.net/forum?id=rywHCPkAW},
}

@article{stable-baselines3,
  author  = {Antonin Raffin and Ashley Hill and Adam Gleave and Anssi Kanervisto and Maximilian Ernestus and Noah Dormann},
  title   = {Stable-Baselines3: Reliable Reinforcement Learning Implementations},
  journal = {Journal of Machine Learning Research},
  year    = {2021},
  volume  = {22},
  number  = {268},
  pages   = {1-8},
  url     = {http://jmlr.org/papers/v22/20-1364.html}
}

@article{castanyer2025stable,
  title={Stable Gradients for Stable Learning at Scale in Deep Reinforcement Learning},
  author={Castanyer, Roger Creus and Obando-Ceron, Johan and Li, Lu and Bacon, Pierre-Luc and Berseth, Glen and Courville, Aaron and Castro, Pablo Samuel},
  journal={arXiv preprint arXiv:2506.15544},
  year={2025}
}

@article{castanyer2025arm,
  title={ARM-FM: Automated Reward Machines via Foundation Models for Compositional Reinforcement Learning},
  author={Castanyer, Roger Creus and Mohamed, Faisal and Castro, Pablo Samuel and Neary, Cyrus and Berseth, Glen},
  journal={arXiv preprint arXiv:2510.14176},
  year={2025}
}

@article{li2017preconditioned,
  title={Preconditioned stochastic gradient descent},
  author={Li, Xi-Lin},
  journal={IEEE transactions on neural networks and learning systems},
  volume={29},
  number={5},
  pages={1454--1466},
  year={2017},
  publisher={IEEE}
}

@inproceedings{ng1999policy,
  title={Policy invariance under reward transformations: Theory and application to reward shaping},
  author={Ng, Andrew Y and Harada, Daishi and Russell, Stuart},
  booktitle={Icml},
  volume={99},
  pages={278--287},
  year={1999},
  organization={Citeseer}
}

@misc{coumans2016pybullet,
  title={Pybullet, a python module for physics simulation for games, robotics and machine learning},
  author={Coumans, Erwin and Bai, Yunfei},
  year={2016}
}
